\documentclass[sigconf, nonacm]{acmart}

\usepackage{amsmath}
  {
      \theoremstyle{plain}
      \newtheorem{assumption}{Assumption}
  }

\usepackage{microtype}
\usepackage{float}

\usepackage{booktabs} 

\usepackage{balance}

\usepackage{epsfig}
\usepackage{epstopdf}

\usepackage{mathtools}
\usepackage{todonotes}

\usepackage[noend]{algorithmic} 
\usepackage{algorithm}

\usepackage{bm}

\usepackage{multirow}

\usepackage{tabularx}
\usepackage{amsfonts}

\usepackage{textcomp}
\usepackage{xcolor}

\usepackage{url}

\usepackage{enumitem}
\setlist[itemize]{noitemsep, topsep=1pt}

\DeclareMathOperator*{\argmax}{arg\,max}
\DeclareMathOperator*{\argmin}{arg\,min}

\newcommand{\Osymbol}{{\mathcal O}}
\newcommand{\BO}[1]{\Osymbol\left(#1\right)}

\newcommand{\E}[1]{\textrm{\bf E}\left[#1\right]}
\renewcommand{\Pr}[1]{\textrm{\bf Pr}\left[#1\right]}

\newcommand{\sgn}{\texttt{sgn}}
\newcommand{\eps}{\varepsilon}

\newcommand{\R} {\mathbb{R}}

\newcommand{\Rd} {\mathbb{R}^d}

\newcommand{\mS} {\mathcal{S}}
\newcommand{\Sd} {\mathcal{S}^{d-1}}

\newcommand{\bX} {\mathbf{X}}
\newcommand{\bR} {\mathbf{R}}

\newcommand{\bH} {\mathbf{H}}
\newcommand{\bD} {\mathbf{D}}

\newcommand{\bw} {\mathbf w}
\newcommand{\bq} {\mathbf q}
\newcommand{\br} {\mathbf r}
\newcommand{\bx} {\mathbf x}
\newcommand{\by} {\mathbf y}

\newcommand{\dotxr}{\bx^\top \br}
\newcommand{\dotyr}{\by^\top \br}
\newcommand{\dotqr}{\bq^\top \br}

\newcommand{\dotxq}{\bx^\top \bq}
\newcommand{\dotyq}{\by^\top \bq}

\newcommand{\Bqr}{B_{\eps}{(\bq)}}
\newcommand{\aBqr}{\widetilde{B}_{\eps}{(\bq)}}
\newcommand{\Br}[1]{B_{\eps}{(#1)}}
\newcommand{\scale}{\sqrt{2 \ln{(D)}}}

\begin{document}

\title{Scalable Density-based Clustering with Random Projections
}

\author{Haochuan Xu}
\email{hxu612@aucklanduni.ac.nz}
\affiliation{%
	\institution{University of Auckland}
	\city{Auckland}
	\country{New Zealand}
}

\author{Ninh Pham}
\email{ninh.pham@auckland.ac.nz}
\affiliation{%
	\institution{University of Auckland}
	\city{Auckland}
	\country{New Zealand}
}

\begin{abstract}

We present \textit{sDBSCAN}, a scalable density-based clustering algorithm in high dimensions with cosine distance. 
Utilizing the neighborhood-preserving property of random projections, sDBSCAN can quickly identify core points and their neighborhoods, the primary hurdle of density-based clustering.
Theoretically, sDBSCAN outputs a clustering structure similar to DBSCAN under mild conditions with high probability. 
To further facilitate sDBSCAN, we present \textit{sOPTICS}, a scalable OPTICS for interactive exploration of the intrinsic clustering structure.
We also extend sDBSCAN and sOPTICS to L2, L1, $\chi^2$, and Jensen-Shannon distances via random kernel features. 
Empirically, sDBSCAN is significantly faster and provides higher accuracy than many other clustering algorithms on real-world million-point data sets.
On these data sets, sDBSCAN and sOPTICS run in a few minutes, while the scikit-learn's counterparts demand several hours or cannot run due to memory constraints. 

\end{abstract}

\maketitle



\section{Introduction}

DBSCAN~\cite{DBSCAN} is one of the most fundamental clustering algorithms with many applications in data mining and machine learning~\cite{ZimekBook}. 
It has been featured in several data analysis tool kits, including scikit-learn in Python, ELKI in Java, and CRAN in R.
In principle, DBSCAN connects neighboring points from nearby high-density areas to form a cluster where the high density is decided by a sufficiently large number of points in the neighborhood.
DBSCAN is parameterized by $(\eps, minPts)$ where $\eps$ is the distance threshold to govern the neighborhood of a point and connect nearby areas, and $minPts$ is the density threshold to identify high-density areas.

Apart from other popular clustering algorithms, including $k$-means variants~\cite{KmeanPlus,KernelKmean} and spectral clustering~\cite{SpectralCluster}, DBSCAN is non-parametric.
It can find the number of clusters, detect arbitrary shapes and sizes, and work on any arbitrary distance measure.

Given a distance measure, DBSCAN has two primary steps, including (1) finding the $\eps$-neighborhood (i.e. points within a radius $\eps$) for every point to discover the density areas surrounding the point and (2) forming the cluster by connecting neighboring points.
The first step is the primary bottleneck as finding $\eps$-neighborhood for every point requires a worst-case $O(n^2)$ time for a data set of $n$ points in high-dimensional space~\cite{DBSCANVisit1,DBSCANVisit2}.
This limits the applications of DBSCAN on modern data sets with millions of points.

Another hurdle of $(\eps, minPts)$-DBSCAN is the choice of $(\eps, minPts)$, which highly depends on the data distribution and distance measure. 
When measuring the clustering accuracy regarding the ground truth (i.e. class labels) using cosine distance, we observe that changing $\eps$ by just 0.005 can diminish the clustering accuracy by 10\% on some data sets.
Therefore, it is essential to not only develop scalable versions of DBSCAN but also to provide feasible tools to guide the parameter setting for such algorithms on large data sets.

\textbf{Prior arts on scaling up DBSCAN.} Due to the quadratic time bottleneck of DBSCAN in high-dimensional space, researchers study efficient solutions to scale up the process of identifying the neighborhood for each point.
These works can be classified into two groups, including balled-based approaches for scaling up \textit{exact} DBSCAN, and index-based and sampling-based approaches for scaling up \textit{approximate} DBSCAN.

Ball-based approaches~\cite{AnyDBC,GapDBC,GroupDbscan} partition the data set into several subsets and extract clusters from these subsets.
These approaches iteratively refine the extracted clustering structures by performing additional $\eps$-neighborhood queries on a small set of important points to obtain the same results as DBSCAN.
However, these approaches still have worst-case quadratic time.

Grid-based indexing~\cite{DBSCANVisit1,GridDbscan} efficiently identifies neighborhood areas by confining the neighbor search exclusively to neighboring grids.
Unfortunately, these approaches only work well on low-dimensional Euclidean space as its complexity grows exponentially to the dimension.
Random projections~\cite{LevelSetDbscan,rpDbscan} have been used to build grid-based or tree-based indexes for faster approximate $\eps$-neighborhoods.
These randomized approaches only work on Euclidean distance and do not offer theoretical guarantees on the clustering accuracy compared to the DBSCAN's result.

Instead of finding the $\eps$-neighborhood for every point, sampling approaches~\cite{DBSCAN++,sngDBSCAN, RoughDbscan} discover the neighborhood for a subset of sampled points and run DBSCAN on this subset to reduce the running time.
DBSCAN++~\cite{DBSCAN++} finds the exact $\eps$-neighborhoods for a subset of random points chosen through either uniform sampling or $k$-center sampling. 
In contrast, sngDBSCAN~\cite{sngDBSCAN} approximates the $\eps$-neighborhood for every point by computing the distance between the point to a subset of random points. 
Though sampling-based DBSCAN variants are simple and efficient, they offer statistical guarantees on the clustering accuracy via level set estimation~\cite{LevelSet} that requires many strong assumptions on the data distribution.
Moreover, selecting suitable parameter values for sampling-based approaches is challenging due to the nature of sampling.

\textbf{Prior arts on selecting parameter values for DBSCAN.} A common approach to select parameter values for $(\eps, minPts)$-DBSCAN is fixing $minPts$ first to smooth the
density estimate~\cite{DBSCANVisit2}.
While $minPts$ is easier to set, $\eps$ is much harder as it depends on the distance measure and data distribution.
OPTICS~\cite{OPTICS} attempts to mitigate the problem of selecting relevant $\eps$ by linearly ordering the data points such that close points become neighbors in the ordering. 
Besides the cluster ordering, OPTICS also computes a reachability distance for each point. 
The dendrogram provided by OPTICS visualizes the density-based clustering results corresponding to a broad range of $\eps$, which indicates a relevant range of $\eps$ for DBSCAN.

Similar to DBSCAN, OPTICS requires the  $\eps$-neighborhood for every point and hence inherits a quadratic time complexity in high-dimensional space.
Even worse, ones often need a large $\eps$ to discover clustering structures on a wide range of $\eps$.
Such large $\eps$ settings need $O(n^2)$ memory as the $\eps$-neighborhood of one point might need $O(n)$ space.
The memory constraint is the primary hurdle that limits the visual tool OPTICS on large data sets. 

\textbf{Contribution.} Inspired by sampling approaches, we observe that the exact $\eps$-neighborhood for every point is not necessary to form and visualize the density-based clustering results.
We leverage the neighborhood-preserving property of random projections~\cite{CEOs} to select high-quality candidates for the discovery of $\eps$-neighborhoods with cosine distance.
Our approach called \textit{sDBSCAN} first builds random projection-based indexing with a sufficiently large number of projection vectors.
The asymptotic property of the concomitant of extreme order statistics~\cite{CEO_Paper} enables us to quickly estimate $\eps$-neighborhoods and form density-based clustering.
Theoretically, sDBSCAN outputs a clustering structure similar to DBSCAN on cosine distance under mild conditions with high probability.
Empirically, sDBSCAN runs significantly faster than other competitive DBSCAN variants while still achieving similar DBSCAN clustering accuracy on million-point data sets.

To further facilitate sDBSCAN, we propose \textit{sOPTICS}, a scalable OPTICS derived from the random projection-based indexing, to guide the parameter setting.
We also extend sDBSCAN and sOPTICS to other popular distance measures, including L2, L1, $\chi^2$, and Jensen-Shannon (JS), that allow random kernel features~\cite{RandomFourier,PAMI12}.

\textbf{Scalability.} Both sDBSCAN and sOPTICS are scalable and multi-thread friendly.
Multi-threading sOPTICS takes \textit{a few minutes} to visualize clustering structures for million-point data sets while the scikit-learn counterpart cannot run due to memory constraints.
Compared to the ground truth on Mnist8m with 8.1 million points, multi-threading sDBSCAN gives 38\% accuracy (measuring by the normalized mutual information NMI) and runs in \textit{15 minutes} on a \textit{single} machine of 2.2GHz 32-core (64 threads) AMD processor with 128GB of DRAM. 
In contrast, kernel $k$-means~\cite{KernelKmean_Nys} achieves 41\% accuracy with Spark, running in \textit{15 minutes} on a \textit{supercomputer} with 32 nodes, each of which has two 2.3GHz 16-core (32 threads) Haswell processors and 128GB of DRAM.

\section{Preliminary}

This section introduces the background of DBSCAN~\cite{DBSCAN,DBSCANVisit2} and its computational challenges in high dimensions.
We also describe OPTICS~\cite{OPTICS}, a DBSCAN-based approach to visualize the clustering structure and select appropriate DBSCAN parameters.
We then elaborate on the connection between range search and approximate nearest neighbor search (ANNS) on executing DBSCAN and OPTICS.
We present a recent advanced random projection method~\cite{CEOs,Falconn++} for ANNS on the extreme setting where the number of random projection vectors is sufficiently large.
The data structures inspired by these approaches can scale up DBSCAN and OPTICS on million-point data sets.

\subsection{DBSCAN}

DBSCAN is a density-based approach that links nearby dense areas to form the cluster.
For a distance measure $dist(\cdot, \cdot)$, DBSCAN has two parameters $\eps$ and $minPts$.
Given the data set $\bX$, for each point $\bq \in \bX$, DBSCAN executes a \textit{range reporting query} $\Bqr$ that finds \textit{all} points $\bx \in \bX$ within the $\eps$-neighborhood of $\bq$, i.e. $\Bqr = \{ \bx \in \bX \, | \, dist(\bx, \bq) \leq \eps \}$.
Based on the size of the range query result, DBSCAN determines $\bq$ as \textit{core} if $|\Bqr| \geq minPts$; otherwise, \textit{non-core} points.
We will call the $\eps$-neighborhood $\Bqr$ as the \textit{neighborhood} of $\bq$ for short.

DBSCAN forms density-based clusters by connecting core points and their neighborhoods where two core points $\bq_1$ and $\bq_2$ are connected if $\bq_1 \in \Br{\bq_2}$ or $\bq_2 \in \Br{\bq_1}$.
A non-core point $\bx$ belonging to a core point $\bq$'s neighborhood (i.e., $\bx \in \Bqr$) will be considered as a border point and will share the same cluster label as $\bq$\footnote{As a non-core point $\bx$ might belong to several core points' neighborhood, DBSCAN uses the label of the first identified core point whose neighborhood contains $\bx$.}.
Non-core points not belonging to any core point's neighborhood will be classified as \textit{noisy} points.

Alternatively, we can see that DBSCAN forms a connected graph $G$ that connects $n$ points together~\cite{DBSCANVisit2,DBSCAN++}.
$G$ will have several connected components corresponding to the cluster structure.
Each connected component of $G$ contains connected core points and their neighborhoods.
Algorithm~\ref{alg:dbscan} shows how DBSCAN works after finding all core points and their neighborhoods.


\begin{algorithm}[!t]
\caption{DBSCAN}
\label{alg:dbscan} 
    \begin{algorithmic}[1] 
    \STATE {\bf Inputs}: $\bX, \eps, minPts$, the set $C = \{ (\bq, \Bqr) \, | \, \bq \text{ is core} \}$
    \STATE $G \gets$ initialize empty graph
    \FOR {each $\bq \in C$}
        \STATE Add an edge (and possibly a vertex or vertices) in $G$ from $\bq$ to all \textit{core} points in $\Bqr$
        \STATE Add an edge (and possibly a vertex) in $G$ from $\bq$ to \textit{non-core} points $\bx \in \Bqr$ if $\bx$ is not connected
    \ENDFOR
    \STATE {\bf return} connected components of $G$
    \end{algorithmic}
\end{algorithm}

\subsection{Challenges of DBSCAN in High Dimensions}

Since we study DBSCAN on $\bX \in \Rd$ of $n$ points, we will elaborate on the challenges of applying DBSCAN to large-scale high-dimensional data sets, including its time complexity and parameter sensitivity.

\textbf{DBSCAN's complexity on L2.}
The running time of DBSCAN is dominated by the cost of executing $n$ range reporting queries $\Bqr$ for $n$ points $\bq \in \bX$, which is $O(n^2)$ for the exact solution.
On low-dimensional Euclidean space, i.e., $d$ is a \textit{constant}, Gan and Tao~\cite{DBSCANVisit1} establish the connection between DBSCAN and the \textit{unit-spherical emptiness checking (USEC)} problem.
This work shows a conditional lower bound $\Omega(n^{4/3})$ for the worst-case running time of DBSCAN based on the notion of Hopcroft \textit{hard}~\cite{Erickson95}.

In high-dimensional Euclidean space where $d$ is large, it is challenging to answer $k$-nearest neighbor ($k$NN) exactly in sublinear time in $n$.
If there exists an algorithm using polynomial indexing space $d^{O(1)}n^{O(1)}$ and solving exact $k$NN in truly sublinear time, the Strong Exponential Time Hypothesis (SETH)~\cite{SETH}, a fundamental conjecture in computational complexity theory, is wrong~\cite{AIR18, Williams14}. 
Hence, we conjecture that solving an \textit{exact} range reporting query in truly sublinear time requires the space exponential in $d$ since we can use $k$NN queries with various $k$ values to answer a range reporting query.
Indeed, several theoretical works~\cite{Range94,Range08,Range10,RangeLSH} seek efficient solutions for answering range reporting queries approximately.
Therefore, solving DBSCAN exactly in subquadratic time in high-dimensional Euclidean space seems hard.  

\textbf{DBSCAN's parameter settings.}
DBSCAN has two parameters $\eps$ and $minPts$ to estimate the empirical density of the point via its neighborhood.
While $minPts$ is easier to set for smoothing the density estimate, the clustering structure provided by DBSCAN is susceptible to $\eps$. 
An incorrect value of $\eps$ will degrade the whole cluster structure.
This is more crucial in high dimensional data sets where the range of $\eps$ is very sensitive.
For instance, when applying DBSCAN with cosine distance on the Pamap2 data set, by varying $\eps$ by just 0.1, there exists $\bq$ such that $\Bqr \approx \bX$, and therefore, the entire data set becomes a single cluster.
To handle this issue, OPTICS~\cite{OPTICS} has been proposed to visualize connected components provided by DBSCAN over a wide range of $\eps$.
Hence, one can select relevant values for $\eps$ to obtain meaningful clustering results.

\subsection{OPTICS}

OPTICS~\cite{OPTICS} attempts to mitigate the problem of selecting relevant $\eps$ by linearly ordering the data points such that close points become neighbors in the ordering. 
For each point $\bx \in \bX$, OPTICS computes a reachability distance from its closest core point.
Each point's cluster ordering and reachability distance are used to construct a reachability-plot dendrogram that visualizes the density-based clustering results corresponding to a broad range of $\eps$.
Valleys in the reachability-plot are considered as clustering indicators.

In principle, given a pair $(\eps, minPts)$, OPTICS first identifies the core points, their neighborhoods, and their core distances.
Then, OPTICS iterates $\bX$, and for each $\bx \in \bX$, computes the smallest reachability distance, called $\bx.reach$, between $\bx$ and the \textit{processed} core points so far.
The point with the minimum reachability distance will be processed first and inserted into the cluster ordering $O$.

The core distance, \textit{coreDist} for short, and reachability distance, \textit{reachDist} for short, are defined as follows.
\begin{align*}
    coreDist(\bq) &=     
    \begin{cases}
        \infty & \text{if $\bq$ is non-core,} \\
        minPts-\text{NN distance} & \text{otherwise.} 
    \end{cases}     
    \\
    reachDist(\bx, \bq) &=     
    \begin{cases}
        \infty & \text{if $\bq$ is non-core,} \\
        max(coreDist(\bq), dist(\bx, \bq)) & \text{otherwise.}        
    \end{cases} 
\end{align*}

For a core point $\bq$, $reachDist(\bx, \bq)$ is $dist(\bx, \bq)$ if $\bx$ is not belonging to $minPts-$NN of $\bq$.
Among several core points whose neighborhood contains $\bx$, OPTICS seeks the \textit{smallest} reachability distance $\bx.reach$ for $\bx$ from these core points.
In other words, $\bx.reach = \min_{\bq_i}{ \left( reachDist(\bx, \bq_i) \, | \, \bq_i \text{ is core and } \bx \in \Br{\bq_i} \right) }$.

OPTICS can be implemented as a nested loop as shown in Algorithm~\ref{alg:optics}.
In the outer loop (Line 4), a random $\bq \in \bX$ is selected and inserted into an empty cluster ordering $O$. 
If $\bq$ is core, each point $\bx \in \Bqr$ is inserted into a \textit{priority queue} $Q$ with $reachDist(\bx, \bq)$ as the key value. 
An inner loop (Line 12) that successively pops $\bq'$ from $Q$ until $Q$ is empty. 
We can see that the priority of $\bq' \in Q$ is determined by their smallest reachability distance w.r.t. current core points processed so far. 
The point with the smallest reachability distance in $Q$ is always popped first (due to the priority queue) and inserted into the ordering $O$. 

We note that our presented OPTICS in Algorithm~\ref{alg:optics} is slightly different from the original version~\cite{OPTICS}.
Since we do not know how to implement decrease-key operation efficiently on the priority queue
~\footnote{C++ STL priority queue does not support decrease-key function.}, we propose a ``lazy deletion'' where we keep adding $\bx$ into $Q$ with the new key.
Though $\bx$ might be duplicated on $Q$, by checking whether or not $\bx$ is processed (Lines 16--18), we can process each point exactly once and output it into the cluster ordering $O$. 
When a point $\bx$ is inserted into the ordering $O$, $\bx.reach$ at the time it was popped out of $Q$ is recorded.
Therefore, by setting $\eps$ large enough, OPTICS outputs a cluster ordering that can be used as a visualization to extract clustering structure for smaller values of $\eps$. 

\textbf{Time and space complexity.} Similar to DBSCAN, the running time of OPTICS is $O(n^2)$ makes it impractical on large-scale data sets.
Fast implementations of OPTICS with large values of $\eps$ will require $O(n^2)$ memory to store the matrix distance between the core points and its neighborhood points.
Such implementations are infeasible for large $n$.

\begin{algorithm}[!t]
\caption{OPTICS}
\label{alg:optics} 
    \begin{algorithmic}[1]
    \STATE {\bf Inputs}: $\bX, \eps, minPts$, the set $C = \{ (\bq, \Bqr, coreDist(\bq), $ \newline $\{ dist(\bx, \bq) \text{ for each } \bx \in \Bqr \} ) \, |  \, \bq \text{ is core} \}$ 
    \STATE Initialize an empty cluster ordering $O$
    \STATE $\bq.reach \gets \infty$ for each $\bq \in \bX$    
    \FOR {each \textit{unprocessed} $\bq \in \bX$}
        \STATE Mark $\bq$ as \textit{processed}, and insert $\bq$ into $O$
        \IF {$\bq$ is core}
            \STATE Seeds $\gets$ empty priority queue $Q$
            \FOR {each $\bx \in \Bqr$}
                \IF {$\bx$ is \textit{unprocessed}}
                    \STATE $\bx.reach \gets min(\bx.reach, reachDist(\bx, \bq))$
                    \STATE Insert $(\bx, \bx.reach)$ into $Q$                    
                \ENDIF
            \ENDFOR
            %
            \WHILE{$Q$ is not empty}
                \STATE $\bq' \gets Q.pop()$
                \IF {$\bq'$ is \textit{processed}}
                    \STATE                \textbf{continue}
                \ENDIF
                \STATE Mark $\bq'$ as \textit{processed}, and insert $\bq'$ into $O$
                \IF {$\bq'$ is core}
                    \FOR {each \textit{unprocessed} $\bx \in B_{\eps}{(\bq')}$}
                       \STATE $\bx.reach \gets min(\bx.reach, reachDist(\bx, \bq'))$                       
                        \STATE Insert $(\bx, \bx.reach)$ into $Q$  
                    \ENDFOR
                \ENDIF
            \ENDWHILE
        \ENDIF        
    \ENDFOR
    \STATE {\bf return} Cluster ordering $O$
    \end{algorithmic}
\end{algorithm}

\subsection{Random projection-based ANNS}

Since the primary bottleneck of DBSCAN is to find core points and their neighborhoods, reducing the running time of this step will significantly improve the performance.
Indeed, most of the techniques in the literature build indexing data structures to speed up the process of finding core points and their neighborhoods.

We will elaborate on recent works~\cite{CEOs,Falconn++} that study extreme order statistics properties of random projection methods for ANNS.
The extreme behavior of random projections is first studied on approximate maximum inner product search, called \textit{CEOs}~\cite{CEOs}.
It is later combined with a cross-polytope LSH family~\cite{Falconn} to improve the ANNS solvers on cosine distance~\cite{Falconn++}.

Given $D$ random vectors $\br_i \in \Rd$, $i \in [D]$, whose coordinates are randomly selected from the standard normal distribution $N(0, 1)$, and  the sign function $\sgn(\cdot)$.
CEOs randomly projects $\bX$ and the query $\bq$ onto these $D$ Gaussian random vectors.
It considers the projection values on specific random vectors that are closest or furthest to $\bq$, e.g., $\argmax_{\br_i}{|\dotqr_i|}$.
Given a sufficiently large $D$, the following lemma states that the projection values on the closest/furthest random vector to $\bq$ preserve the inner product approximately.
\begin{lemma}~\cite{CEOs}\label{lm:CEOs}
	For two points $\bx, \bq \in \Sd$ and significantly large $D$ random vectors $\br_i$, w.l.o.g. we assume that $\br_1 = \argmax_{\br_i}{|\dotqr_i|}$. Then, we have
	\begin{align}\label{eq:CEOs}
	\dotxr_1 \sim N \left( \mathtt{sgn}(\dotqr_1) \cdot \dotxq \,  \sqrt{2 \ln{(D)}} \, , 1 - (\dotxq)^2\right) \, .
	\end{align}	
\end{lemma} 

As a geometric intuition, we can see that, for significantly large $D$ random vectors, if $\br_1$ is closest or furthest to $\bq$, the projection values of all points in $\bX$ onto $\br_1$ tend to preserve their inner product order with $\bq$.
Importantly, given a constant $k > 0$, Lemma~\ref{lm:CEOs} also holds for the top-$k$ closest/furthest vectors to $\bq$ due to the asymptotic property of extreme normal order statistics~\cite{CEO_Paper, CEOs}.
Therefore, by maintaining a few points that are closest/furthest to random vectors, we can answer ANNS accurately and efficiently.
In the next section, we will utilize this observation to significantly reduce the cost of identifying core points and, hence, the running time of DBSCAN and OPTICS in high dimensions.

\begin{algorithm}[!t]
	\caption{Preprocessing}
	\label{alg:preprocessing}	
	\begin{algorithmic} [1]
            \STATE {\bf Inputs}: $\bX \subset \Sd$, $D$ random vectors $\br_i$, $k, m = O(minPts)$		
            \STATE \textbf{for} each $\bq \in \bX$, compute and store top-$k$ {\color{blue}closest} and top-$k$ {\color{blue}furthest} vectors $\br_i$ to $\bq$.
            \STATE \textbf{for} each random vector $\br_i$, compute and store top-$m$ {\color{blue}closest} and top-$m$ {\color{blue}furthest} points to $\br_i$.		
		\normalsize		
	\end{algorithmic}
\end{algorithm}

\section{Random projection-based methods}

While previous works leverage the extreme behavior of random projections to improve ANNS solvers, we utilize it to find the core points and their neighborhoods.
We first present novel scalable DBSCAN and OPTICS versions, called \textit{sDBSCAN} and \textit{sOPTICS}, with the \textit{cosine distance} on the unit sphere.
We then leverage well-known random feature embeddings~\cite{RandomFourier, PAMI12} to extend our proposed \textit{sDBSCAN} and \textit{sOPTICS} to other popular distance measures, including L1 and L2 metrics, and widely used similarity measures for image data, including $\chi^2$ and Jensen-Shannon (JS).

\subsection{Preprocessing and finding core points}

The primary execution of DBSCAN is to identify core points and their neighborhoods to form connected cluster components.
In particular, $(\eps, minPts)$-DBSCAN identifies $\bq \in \bX$ as a core point if $|\Bqr| \geq minPts$.
We present how random projections can speed up the identification of core points and approximate their neighborhoods with cosine distance.

For each point $\bq \in \bX \subset \mS^{d-1}$, we compute $D$ random projection values $\dotqr_i$ where $i \in [D]$. 
W.l.o.g., we assume that $\br_1 = \argmax_{\br_i}{|\dotqr_i|} = \argmax_{\br_i}{\dotqr_i}$.
Lemma~\ref{lm:CEOs} indicates that, given a sufficiently large $D$, points closer to $\br_1$ tend to have smaller distances to $\bq$.
Hence, computing the distance between $\bq$ to the top-$m$ points closest to $\br_1$ where $m \geq minPts$ suffices to ensure whether or not $\bq$ is a core point.

As Lemma~\ref{lm:CEOs}  holds for the top-$k$ closest and top-$k$ furthest random vectors to $\bq$, for each random vector $\br_i$, we maintain the top-$m$ closest points and the top-$m$ furthest points to $\br_i$, respectively.
Given a point $\bq$, we compute the distances between $\bq$ to the top-$m$ closest/furthest points corresponding to the top-$k$ closest/furthest random vectors of $\bq$.
By considering these $2k$ specific vectors and computing $2km$ distances, we can improve the accuracy of identifying the core points and enrich their neighborhoods.

\textbf{Preprocessing.} 
Algorithm~\ref{alg:preprocessing} shows how we preprocess the data set.
For each point $\bq \in \bX$, we keep top-$k$ closest/furthest random vectors.
For each random vector $\br_i$, we keep top-$m$ closest/furthest points.
In practice, setting $m = O(minPts)$ suffices to identify core points and approximate their neighborhoods to ensure the quality of sDBSCAN and sOPTICS.
In theory, we need to adjust $m$ depending on the data distribution to guarantee the clustering quality.

\begin{algorithm}[!t]
	\caption{Finding core points and their neighborhoods}
	\label{alg:neighborhood} 									
	\begin{algorithmic} [1]
            \STATE {\bf Inputs}: $\bX \subset \Sd$, $D$ random vectors $\br_i$, $k, \eps, m = O(minPts)$		
            \STATE Initialize an empty set $\aBqr$ for each $\bq \in \bX$
            \FOR{\text{each} $\bq \in \bX$}
                \FOR{\text{each} $\br_i$ from top-$k$ {\color{blue}closest} random vectors of $\bq$}
                    \FOR{\text{each} $\bx$ from top-$m$ {\color{blue}closest} points of $\br_i$}
                        \IF{$dist(\bx, \bq) \leq \eps$}
                            \STATE Insert $\bx$ into $\aBqr$ and insert $\bq$ into $\widetilde{B}_{\eps}(\bx)$
                        \ENDIF                        
                    \ENDFOR                    
                \ENDFOR
                \FOR{\text{each} $\br_i$ from top-$k$ {\color{blue}furthest} random vectors of $\bq$}
                    \FOR{\text{each} $\bx$ from top-$m$ {\color{blue}furthest} points of $\br_i$}
                        \IF{$dist(\bx, \bq) \leq \eps$}
                            \STATE Insert $\bx$ into $\aBqr$ and insert $\bq$ into $\widetilde{B}_{\eps}(\bx)$
                        \ENDIF                        
                    \ENDFOR                    
                \ENDFOR
            \ENDFOR
            %
            \FOR{\text{each} $\bq \in \bX$}
                \IF{$|\aBqr| \geq minPts$}
                    \STATE Identify $\bq$ as the core point
                    \STATE Store $\aBqr$ as an approximate result of $\Bqr$
                    \STATE Store $dist(\bx, \bq)$ for each $\bx \in \aBqr$  \COMMENT{\textit{Only for sOPTICS}}
                \ENDIF
            \ENDFOR            
		\normalsize		
	\end{algorithmic}
\end{algorithm}

\textbf{Finding core points and their neighborhoods.} 
After preprocessing, we run Algorithm~\ref{alg:neighborhood} to identify core points and approximate their neighborhoods using $D$ random vectors.
Given $m = O(minPts)$, for each point $\bq \in \bX$, we only compute the distances to the $O(k \cdot minPts)$ points~\footnote{Due to duplicates, $2k \cdot m$ is the maximum number of distances for each point.} from the top-$k$ closest/furthest random vectors to $\bq$.
For a small constant $k$, we only need $O(d \cdot minPts)$ time, compared to $O(dn)$ of the exact solution, to identify a core point $\bq$ and its neighborhood subset $\aBqr \subseteq \Bqr$.
Though we might miss a few core points and points in their neighborhoods, our theoretical analysis shows that the identified core points and their neighborhood subsets can recover the DBSCAN's clustering structure under common assumptions.

\subsection{sDBSCAN and sOPTICS}

After identifying core points and their approximate neighborhoods, we form the density-based clustering, as shown in Algorithm~\ref{alg:sdbscan}.
Two core points $\bq_1, \bq_2$ are connected to each other if $\bq_1 \in \widetilde{B}_{\eps}(\bq_2)$ or $\bq_2 \in \widetilde{B}_{\eps}(\bq_1)$.
Though $\aBqr \subseteq \Bqr$, we will show that sDBSCAN can output the DBSCAN's result under common assumptions.

To guide DBSCAN's parameter setting, Algorithm~\ref{alg:soptics} presents \textit{sOPTICS}, an OPTICS variant with the identified core points and their neighborhood subsets provided by Algorithm~\ref{alg:neighborhood}.
Given the identified core point $\bq$ and $\aBqr$, we store the set $\{ dist(\bx ,\bq) \text{ for each } \bx \in \aBqr \}$ (Line 16 of Alg.~\ref{alg:neighborhood}).
Using this set, we can estimate $coreDist(\bq)$ as the $minPts$-NN distance between $\bq$ and $\aBqr$.
Though such estimation is an \textit{upper bound} of $coreDist(\bq)$, the tightness of the upper bound and the accurate neighborhood approximation make the reachability-plot provided by sOPTICS nearly identical to the original OPTICS with the same valley areas.
Therefore, sOPTICS is a fast and reliable tool to guide DBSCAN and sDBSCAN parameter settings and to visualize the clustering structure.

\begin{algorithm}[!t]
	\caption{sDBSCAN}
	\label{alg:sdbscan} 									
	\begin{algorithmic} [1]
		\STATE {\bf Inputs}: $\bX \subset \Sd$, $D$ random vectors $\br_i$, $\eps$, $minPts$	
            \STATE {Call Algorithm~\ref{alg:preprocessing} for preprocessing with $m = O(minPts)$}
            \STATE {Call Algorithm~\ref{alg:neighborhood} to find the set \newline $C = \{ (\bq, \aBqr) \, | \, \bq \text{ is identified as core} \}$}
            \STATE {Call DBSCAN (Alg.~\ref{alg:dbscan}) given the output $C$ from Algorithm~\ref{alg:neighborhood}}
		\normalsize		
	\end{algorithmic}
\end{algorithm}

\textbf{Multi-threading.}
Like DBSCAN and OPTICS, the main computational bottlenecks of both sDBSCAN and sOPTICS are identifying core points and approximating their neighborhood. 
Fortunately, Algorithm~\ref{alg:preprocessing} and~\ref{alg:neighborhood} are fast and parallel-friendly. 
We only need to add \textbf{\texttt{\#pragma omp parallel}} directive on the \textbf{\texttt{for}} loops to run in multi-threads.
This enables sDBSCAN and sOPTICS to deal with million-point data sets in \textit{minutes} while the scikit-learn counterparts need hours or even cannot finish.

\section{Theoretical analysis}

We first prove sDBSCAN can identify any core point with high probability.
We then show how to set $m$ adaptively to the data distribution to ensure sDBSCAN recovers the clustering result of DBSCAN under common assumptions.
We also discuss practical settings and heuristic improvements for sDBSCAN and sOPTICS.

\subsection{Identify core points}

For each point $\bx \in \bX \subset \mS^{d-1}$, we compute $D$ random projection values $\dotxr_i$ where $i \in [D]$. These random projection values are used to identify the core points in $\bX$ and approximate their neighborhoods.
For simplicity, we analyze a single point $\bq \in \bX$ with the cosine distance.

For a sufficiently large $D$, Lemma~\ref{lm:CEOs} holds for top-$k$ closest/furthest random vectors.
We assume that $\br_1 = \argmax_{\br_i}{|\dotqr_i|} = \argmax_{\br_i}{\dotqr_i}$ w.l.o.g..
We denote by $\bx \in \Bqr$ and $\by \in \bX \setminus \Bqr$ any close and far away points to $\bq$ regarding the distance threshold $\eps$.
Defining random variables $X = \dotxr_1, Y = \dotyr_1$, from Eq.(\ref{eq:CEOs}), we have
\begin{align*}
    X \sim N \left( \dotxq \,  \scale \, , 1 - (\dotxq)^2\right) \, , \, Y \sim N \left( \dotyq \,  \scale \, , 1 - (\dotyq)^2\right) 
\end{align*}

Let $\alpha_{\bx} = \left(\bx^{\top}\bq - \dotyq \right) / \left( \sqrt{1 - \left( \bx^{\top}\bq \right)^2} + \sqrt{1  - \left( \dotyq \right)^2} \right)$.
From the Chernoff bound~\cite{Chernoff} for the Gaussian variable $Y - X$, we have
\begin{align}\label{eq:CEOs-Chernoff}
\Pr{Y \geq X} &\leq D^{-\left(\dotxq - \dotyq\right)^2 / \left( \sqrt{1 - \left( \dotxq \right)^2} + \sqrt{1  - \left( \dotyq \right)^2} \right)^2} = D^{-\alpha_{\bx}^2} .
\end{align}
We can see that the order of projection values onto $\br_1$ between $\bx$ and $\by$, i.e., $\dotxr_1 \geq \dotyr_1$, preserves their inner product order, i.e., $\dotxq \geq \dotyq$, with high probability when $D$ is sufficiently large.
To estimate $\Bqr$, we sort $n$ projection values onto $\br_1$, and consider the top-$m$ points closest to $\br_1$ as the top-$m$ candidates where $m \geq minPts$.
Among these $m$ candidates, if $minPts$ points are within $\eps$ to $\bq$, we safely conclude that $\bq$ is the core point.
The following lemma justifies that $m = minPts$ suffices to identify any core point.

\begin{lemma}\label{lm:core}
    Given a sufficiently large $D$, for a given core point $\bq \in \bX \subset \Sd$ where $|\Bqr| \geq minPts$ and its closest random vector $\br_1$, the top-$minPts$ points with largest projection values onto $\br_1$ are a subset of $\Bqr$ with probability at least $1 - 1/n$.
\end{lemma}

\begin{algorithm}[!t]
	\caption{sOPTICS}
	\label{alg:soptics}
	\begin{algorithmic} [1]
		\STATE {\bf Inputs}: $\bX \subset \Sd$, $D$ random vectors $\br_i$, $\eps$, $minPts$	
            \STATE {Call Algorithm~\ref{alg:preprocessing} for preprocessing with $m = O(minPts)$}
            \STATE {Call Algorithm~\ref{alg:neighborhood} to find the set $C= \{ (\bq, \aBqr, \newline \{ dist(\bx, \bq) \text{ for each } \bx \in \aBqr \}) \, | \, \bq \text{ is identified as core} \}$}
            \STATE {Use the $minPts-$NN distance between $\bq$ and $\bx \in \aBqr$ as $coreDist(\bq)$ for each identified core point $\bq$}
            \STATE {Call OPTICS (Alg.~\ref{alg:optics}) given the output $C$ from Algorithm~\ref{alg:neighborhood}}
		\normalsize		
	\end{algorithmic}
\end{algorithm}

\begin{proof}
We consider the $n$ projection values onto $\br_1$, i.e. $\bX^{\top} \br_1$, in the descending order. 
For a given point $\by \in \bX \setminus \Bqr$, let $Y = \dotyr_1$.
For any $\bx_i \in \Bqr$, we define random variables $X_i = \bx_i^\top \br_1$. 
Denote by $\overline{\alpha} = \left( \sum_{\bx_i \in \Bqr}{\alpha_{\bx_i}} \right) / |\Bqr|$ the average value for all $\alpha_{\bx_i}$ for $\bx_i \in \Bqr$ and $\by$, by applying Eq.~(\ref{eq:CEOs-Chernoff}) we have
\begin{align*}
    &\Pr{\by \text{ is ranked as top-1}} \leq \prod_{\bx_i \in \Bqr}{\Pr{Y \geq X_i}} \,  \\
    &\leq D^{-\sum_{\bx_i \in \Bqr}{\alpha^2_{\bx_i}}} \leq D^{- |\Bqr| \overline{\alpha}^{\, 2}} \, .
\end{align*}
Let $B_{minPts}(\bq)$ be the set of top-$minPts$ nearest neighbors of $\bq$ in $\bX$.
We define $\overline{\alpha}_* = \left( \sum_{\bx_i \in \Bqr \setminus B_{minPts}(\bq)}{\alpha_{\bx_i}} \right) / \left( |\Bqr| - minPts \right)$.
Using the same argument, we bound the probability that $\by$ is ranked on the top-$minPts$ points among $n$ projection values onto $\br_1$ as
$$
    \Pr{\by \text{ is ranked on top-}minPts} \leq D^{- \left( |\Bqr| - minPts \right)\overline{\alpha}_*^{\, 2}} \, .
$$
By setting $D = n^{2 / \left( |\Bqr| - minPts \right)\overline{\alpha}_*^{\, 2}}$, this probability is bounded by $1/n^2$.


For any $\by_j \in \bX \setminus \Bqr$, let $\alpha_* = \argmin_{\by_j \in \bX \setminus \Bqr} \overline{\alpha}_*$ and $D = n^{2 / \left( |\Bqr| - minPts \right)\alpha_*^2}$,
applying the union bound, we have
\begin{align*}
    &\Pr{\text{Top-}minPts \text{ points closest to $\br_1$ are in } \Bqr} \, \\
    &\geq 1 - n \cdot \Pr{\by_j \text{ is ranked among the top-}minPts \text{ positions} } \, \\
    &\geq 1 - 1/n \, .  
\end{align*} 

Hence, we prove the claim given a sufficiently large $D$.
\end{proof}
Lemma~\ref{lm:core} indicates that setting $m = minPts$ suffice to identify whether a point $\bq$ is core.
As Lemma~\ref{lm:CEOs} holds with the top-$k$ closest/furthest random vectors to $\bq$ given a constant $k$, so does Lemma~\ref{lm:core}.
Therefore, Algorithm~\ref{alg:neighborhood} (Lines 4--11) runs the same procedure on these $2k$ specific random vectors, increasing the core point identification's accuracy and enriching the core point's neighborhood.
This is essential to ensure the quality of sDBSCAN for clustering and its corresponding sOPTICS for visualization.

We note that Lemma~\ref{lm:core} holds given significantly large $D$ random vectors, i.e. $D = n^{2 / \left( |\Bqr| - minPts \right)\alpha_*^2}$.
Empirically, we observe that, for any core point $\bq$ regarding $(\eps, minPts)$, $|\Bqr|$ is often substantially larger than $minPts$.
Furthermore, the distance gap between $\bq$ and points inside and outside $\Bqr$ is significant, making $\alpha_*^2$ large.
Hence, setting $D = 1,024$ suffices to achieve the clustering quality on many real-world data sets.

\subsection{Ensure sDBSCAN's quality}

Ensuring the clustering quality without any assumptions about the data distribution is hard.
Following up on previous works~\cite{DBSCAN++,sngDBSCAN}, we describe the common assumption used in analyzing the performance of sDBSCAN given the approximate neighborhood of identified core points.

\begin{assumption}
  There exists a constant $t$ such that, for any pair of nearby core points $\bq_1, \bq_2$, if $dist(\bq_1, \bq_2) \leq \eps$, then $B_{\eps}(\bq_1)$ and $B_{\eps}(\bq_2)$ share at least $t$ common core points.
\end{assumption}

Since two connected core points share at least $t$ common core points, this assumption ensures that the density-based clusters would not become arbitrarily thin anywhere.
Otherwise, 
it is a good indicator that we need to select different parameter settings $(\eps, minPts)$ to reflect the clustering structure.
Our assumption is similar to the third assumption in~\cite{sngDBSCAN}, and our proof technique is inspired by~\cite{sngDBSCAN}.
However, we do not need the other two assumptions in~\cite{sngDBSCAN} since we can leverage the distance preservation property of CEOs.

Given that DBSCAN produces $l$ density-based clusters $C_1, C_2, \ldots, C_l$, it can be seen easily that $t$ will be the lower bound of the min-cut of the subgraph $G_i$ formed by connecting all core points in the cluster $C_i$ together.
We will use the following lemma~\cite{sngDBSCAN,Karger99} to guarantee the quality of sDBSCAN.
\begin{lemma}~\label{lm:graph}
    Let $G$ be a graph of $n$ vertices with min-cut $t$ and $0 < \delta < 1$.
    There exists a universal constant $c$ such that if $s \geq \frac{c \left( \log{(1/\delta)} + \log{(n)}\right)}{t}$, 
    then with probability at least $1 - \delta$, the graph $G_s$ derived by sampling each edge of $G$ with probability $s$ is connected.
\end{lemma}

Now, the rest of the analysis shows that for a given core point $\bq \in G_i$, we guarantee that sDBSCAN will find any point $\bx \in \Bqr$ with probability at least $c \left ( \log{(1/\delta)} + \log{(|G_i|)} \right ) / t$.

Consider if $dist(\bx, \bq) \leq \eps$ then $\dotxq \geq 1 - \eps^2/2$, we make a change in our algorithm.
For each random vector $\br_i$, we maintain two sets $S_i = \{ \bx \in \bX \, | \, \dotxr_i \geq (1 - \eps^2/2) \scale \}$ and $R_i = \{ \bx \in \bX \, | \, \dotxr_i \leq -(1 - \eps^2/2) \scale \}$, instead of keeping only top-$m$ closest/furthest points.
In other words, for each random vector $\br_i$, the value of $m$ is set \textit{adaptively} to the data distributed around $\br_i$.
Now, given a point $\bq$ and its closest random vector $\br_1$, we compute $dist(\bx, \bq)$ for all $\bx \in S_1$ where $S_1 = \{ \bx \in \bX \, | \, \dotxr_1 \geq (1 - \eps^2/2) \scale \}$.
By checking all $\bx \in S_1$, we guarantee to select any $\bx \in \Bqr$ with a probability at least 1/2 due to the Gaussian distribution in Eq. (\ref{eq:CEOs}).
%

For any point $\bq$, by computing the distance between $\bq$ and all points in $S_i$ or $R_i$ corresponding to the top-$k$ closest/furthest random vectors of $\bq$, we can boost the probability to identify any $\bx \in \Bqr$ to at least $1 - \left(1/2\right)^{2k}$ due to the asymptotic independence among these $2k$ vectors~\cite{CEOs,CEO_Paper}.
Therefore, by Lemma~\ref{lm:graph}, if $t \left( 1 - \left(1/2\right)^{2k} \right) \geq c \left ( \log{(1/\delta)} + \log{(n_i)} \right )$ where $n_i$ is the number of core points of $C_i$, we can guarantee that the cluster structures provided by DBSCAN and sDBSCAN are the same with probability at least $1 - \delta$.
We are now ready to state our main result.

\begin{lemma}\label{lm:main}
    Let $G_1, G_2, \ldots, G_l$ be connected subgraphs produced by DBSCAN where each $G_i$ corresponds to a cluster $C_i$ with $n_i$ core points. 
    Assume that any pair of nearby core points $\bq_1, \bq_2$, if $dist(\bq_1, \bq_2) \leq \eps$, then $B_{\eps}(\bq_1)$ and $B_{\eps}(\bq_2)$ share at least $t$ common core points.
    There exists a constant $c$ such that if $t \left( 1 - \left(1/2\right)^{2k} \right) \geq c \left ( \log{(1/\delta)} + \log{(n_i)} \right )$ for $i \in [l]$, sDBSCAN will recover $G_1, G_2, \ldots, G_l$ as clusters with probability at least $1 - \delta$.
\end{lemma}

Lemma~\ref{lm:main} indicates that when the cluster $C_i$ of size $n_i$ is not thin anywhere, i.e. the neighborhoods of any two connected core points share $t = O(\log{(n_i)})$ common core points, sDBSCAN can recover $C_i$ with probability $1 - 1/n_i$. 
However, it comes with the cost of maintaining larger neighborhoods around the random vector $\br_i$ (i.e. $S_i$, $R_i$) and hence causes significant computational resources.

In practice, a core point $\bq$ is often surrounded by many other core points in a dense cluster.
Therefore, instead of maintaining the sets $S_i, R_i$ for each random vector $\br_i$, we can keep the top-$m$ points of these sets where $m = O(minPts)$.
This setting substantially reduces the memory usage and running time of sDBSCAN without degrading clustering results.


\subsection{Discuss sOPTICS's quality}


Ensuring sOPTICS recovers OPTICS's result is difficult without any further assumptions as OPTICS's result is sensitive to the order of processed core points and their exact neighborhoods.
Hence, we will discuss sOPTICS's quality in practical scenarios.

Like sDBSCAN, sOPTICS only considers top-$m$ closest/furthest points to any random vector where $m = O(minPts)$.
Given a core point $\bq$, the core distance of $\bq$ derived from the set $\aBqr$ is an \textit{upper bound} of $coreDist(\bq)$.
Since a core point $\bq$ is often surrounded by many other core points in a dense cluster, $\aBqr$ tends to contain 
core points.
Hence, the upper bound of reachability distance provided by sOPTICS is tight.
When the clustering structure is well separated, i.e., valleys in the reachability-plot are deep, sOPTICS with $m = O(minPts)$ is a reliable tool to guide the selection of $(\eps, minPts)$ for DBSCAN.
Importantly, the extra space of sOPTICS is linear, i.e. $O(nk \cdot minPts)$, as the approximate neighborhood size of each point is $O(k \cdot minPts)$.

\subsection{Extend to other distance measures}

While previous subsections present and analyze sDBSCAN and sOPTICS on cosine distance, this subsection will extend them to other popular distance measures, including L2, L1, $\chi^2$, and Jensen-Shannon (JS).
We will utilize the random features~\cite{RandomFourier, PAMI12} to embed these distances into cosine distance.
In particular, we study fast randomized feature mapping $f: \R^d \mapsto \R^{d'}$ such that $\E{ f(\bx)^{\top} f(\bq) } = K(\bx, \bq)$ where $K$ is the kernel function.
We study Gaussian, Laplacian, $\chi^2$, and JS kernels as their randomized mappings are well-studied and efficiently computed.
Also, the embeddings' extra costs are negligible compared to those of sDBSCAN and sOPTICS.

Given $\bx = \{x_1, \ldots , x_d \}, \by = \{y_1, \ldots , y_d \}$ and $\sigma > 0$, the following are the definitions of the investigated kernels.
\begin{align*}
    K_{L}(\bx, \by) &= e^{-\| \bx - \by\|_1 / \sigma} \, ; \, K_{G}(\bx, \by) = e^{-\| \bx - \by\|_2^2 / 2\sigma^2} \, ; \\
    K_{\chi^2}(\bx, \by) &= \sum_i \frac{2x_i y_i}{x_i + y_i} \, ; \\
    K_{JS}(\bx, \by) &= \sum_i \frac{x_i}{2}\log{\left( \frac{x_i + y_i}{x_i} \right)} + \frac{y_i}{2}\log{\left( \frac{x_i + y_i}{y_i} \right)} \, .
\end{align*}
%

Due to the space limit, we present the random Fourier embeddings~\cite{RandomFourier} for $K_L$ and $K_G$ with $\sigma = 1$.
We first generate $d'$ random vectors $\bw_i, i \in [d']$ whose coordinates are from $N(0, 1)$ for $K_G$ and $Lap(0, 1)$ for $K_L$.
Our randomized mappings are: 
$$f(\bx) = \frac{1}{\sqrt{d'}}\{ \sin{(\bw_i^{\top} \bx)}, \cos{(\bw_i^{\top} \bx)} \, | \, i \in [d'] \} \in \mS^{2d' - 1} \,  .$$
Since $|f(\bx)^{\top} f(\by)| \leq \| f(\bx)\|_2 \| f(\by)\|_2 = 1$, the Hoeffding’s inequality guarantees:
$\Pr{ |f(\bx)^{\top} f(\by) - K(\bx, \by)| \geq \delta } \leq e^{-d' \cdot \, \delta^2 / 2} \, .$
By selecting $d' = O(\log{(n)})$, the randomized mapping $f$ preserves well the kernel function of every pair of points.
Hence, sDBSCAN and sOPTICS on these randomized embeddings output similar results to DBSCAN and OPTICS on the corresponding distance measures.

\textbf{Complexity.}
For $K_G, K_L$, the embedding time is $O(dd')$.
For $K_{\chi^2}, K_{JS}$, the embedding time is $O(d')$ for $d'$ random features by applying the sampling and scaling approaches~\cite{PAMI12}.
Empirically, the random feature construction time is negligible compared to the sDBSCAN and sOPTICS time.
As we execute random projections on the constructed random features for each point and compute $dist(\bx, \by)$ using the original data, we only need a small extra space to store random vectors $\bw_i$.


\subsection{From theory to practice}\label{sec:practice}

We observe that the core point $\bq$ often falls into the high-density area where $\Bqr$ tends to be dominated by core points in real-world data sets.
Hence, $m = O(minPts)$ suffices to identify core points and connect them to form the density-based clustering.

We note that sDBSCAN requires $D$ random vectors whose coordinates are randomly selected from $N(0, 1)$, which requires $O(dD)$ time to process a point.
This work will use the Structured Spinners~\cite{Falconn} that exploits Fast Hadamard Transform (FHT) to reduce the cost of random projections.
In particular, we generate 3 random diagonal matrices $\bD_1, \bD_2, \bD_3$ whose values are randomly chosen in $\{+1, -1\}$.
The random projection $\bR \bx$ is simulated by the mapping $\bx \mapsto \bH \bD_3 \bH \bD_2 \bH \bD_1 \bx$ where $\bH$ is the Hadamard matrix.
With FHT, the random projection can be simulated in $\BO{D \log{(D)}}$ time and use additional $O(D)$ extra space to store random matrices.

\begin{figure*} [ht]
	\centering
\includegraphics[width=1\textwidth]{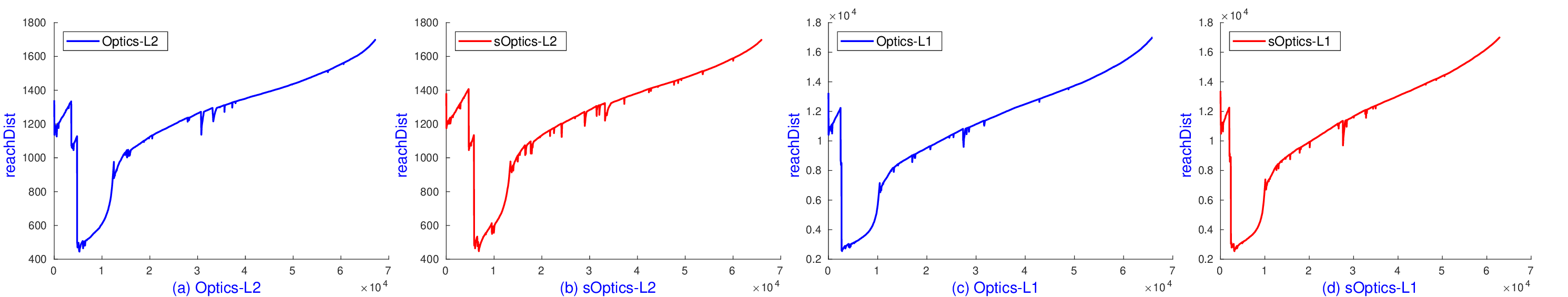}
	\caption{Reachability-plot dendrograms of OPTICS and sOPTICS over L2 and L1 on Mnist. While sOPTICS needs less than \textit{30 seconds}, scikit-learn OPTICS requires \textit{1.5 hours} on L2 and \textit{0.5 hours} on L1.}
	\label{fig:Optic-Mnist}
\end{figure*}

\textbf{A random projection-based heuristic to cluster non-core points.}
As sDBSCAN focuses on identifying core points and approximating their neighborhoods, it misclassifies border and noisy points.
As the neighborhood size of detected core points is upper bound by $2km$, 
sDBSCAN might suffer accuracy loss on data sets with a significantly large number of border and noisy points.
Denote by $C$ and $\overline{C}$ the set of identified core and non-core points by sDBSCAN, we propose a simple heuristic to classify any $\bq \in \overline{C}$.
We will build a nearest neighbor classifier (1NN) where training points are \textit{sampled} from $C$, and scale up this classifier with the CEOs approach~\cite[Alg. 1]{CEOs}.
We call sDBSCAN with the approximate 1NN heuristic as \textit{sDBSCAN-1NN}.

In particular, for each sampled core point $\bx \in C$, we recompute their random projection values with the stored $D$ random vector $\br_i$ as we do not keep these values after preprocessing.
For each non-core point $\bq \in \overline{C}$, we retrieve the \textit{precomputed} top-$k$ closest/furthest vectors and use the projection values of $\bx$ at these top-$k$ vectors to estimate $\dotxq$.
To ensure this heuristic does not affect sDBSCAN time where $|C| \approx |\overline{C}| \approx n/2$, we sample $0.01 n$ core points from $C$ to build the training set.
Empirically, sDBSCAN time dominates this heuristic time, and the extra space usage to store $0.01 n D$ projection values is negligible.



\subsection{Time and space complexity} 

We first analyze the time complexity of the preprocessing and finding neighborhood steps with $m = O(minPts)$.
Then we discuss the time and space complexity of sDBSCAN and sOPTICS.

\textbf{Time complexity.} Using FHT, the preprocessing (Alg.~\ref{alg:preprocessing}) runs in $O \left( n D \log{(D)} + nD\log{(k)} + nD\log{(minPts)}) \right)$ time since we retrieve top-$k$ closest/furthest vectors for each $\bq \in \bX$, and top-$m$ closest/furthest points for each random vector $\br_i$. 
Finding the neighborhood for $n$ points (Alg.~\ref{alg:neighborhood}) runs in $O(dnk \cdot minPts)$ time since each point needs $O(k \cdot minPts)$ distance computations.
For a constant $k$, sDBSCAN runs in $O(dn \cdot minPts + nD \log{(D)})$ time.

For each point $\bq$, sOPTICS keeps $|\aBqr| = O(k \cdot minPts)$ points and distance values.
The size of the priority queue $Q$ of sOPTICS is $O(nk \cdot minPts)$.
Therefore, for a constant $k$, sOPTICS runs in $O(dn \cdot minPts + nD \log{(D)} + n \cdot minPts \cdot \log{(n \cdot minPts)})$.
When $D = o(n)$, both sDBSCAN and sOPTICS run in \textit{subquadratic} time.

Empirically, the cost of the preprocessing step $O(nD \log{(D)})$ is much smaller than the cost of finding neighborhood $O(dnk \cdot minPts)$ in large data sets.
This is because finding the neighborhood for each point requires $O(k \cdot minPts)$ distance computations that need expensive random memory access, while the cache-friendly FHT uses sequential memory access.
Fortunately, preprocessing and finding neighborhood steps are elementary to run in parallel.
Our multi-threading implementation of sDBSCAN and sOPTICS shows a 10$\times$ speedup with 64 threads.

\textbf{Space complexity.} 
Our sDBSCAN needs $O(nk + D \cdot minPts)$ extra space to store $O(k)$ closest/furthest vectors for each point, and $O(minPts)$ points closest/furthest to each random vector.
When $D = o(n)$, sDBSCAN's additional memory is negligible compared to the data size $O(nd)$.
Due to the priority queue of size $O(nk \cdot minPts)$, sOPTICS uses $O(nk + D \cdot minPts + nk \cdot minPts)$ extra space, which is linear when $D = o(n)$, and \textit{independent} on any value of $\eps$.

\section{Experiment}

We implement sDBSCAN and sOPTICS in C++ and compile with \texttt{g++ -O3 -std=c++17 -fopenmp -march=native}.
We conducted experiments on Ubuntu 20.04.4 with an AMD Ryzen Threadripper 3970X 2.2GHz 32-core processor (64 threads) with 128GB of DRAM. 
We present empirical evaluations on the clustering quality compared to the DBSCAN's clustering result and  the ground truth (i.e. data labels) to verify
our claims, including:
\begin{itemize}
	\item sDBSCAN and sOPTICS can output the same results as DBSCAN and OPTICS using significantly smaller computational resources on high-dimensional data sets.
	\item sDBSCAN with the suggested parameter values provided by sOPTICS runs significantly faster and achieves competitive accuracy compared to other clustering algorithms. 
 \item Multi-threading sDBSCAN and sOPTICS run in \textit{minutes}, while the scikit-learn counterparts cannot run on million-point data sets.
\end{itemize}

Our competitors include pDBSCAN~\cite{DBSCANVisit1}~\footnote{\url{https://sites.google.com/view/approxdbscan}} as a representative grid-based approach, DBSCAN++~\cite{DBSCAN++}~\footnote{\url{https://github.com/jenniferjang/dbscanpp}} and sngDBSCAN~\cite{sngDBSCAN}~\footnote{\url{https://github.com/jenniferjang/subsampled_neighborhood_graph_dbscan}} as representatives for sampling-based approaches.
DBSCAN++ has two variants, including DBSCAN++ with uniform initialization (called uDBSCAN++) that uses KD-Trees to speed up the search of core points and $k$-center initialization (called kDBSCAN++).
We also compare with multi-threading scikit-learn implementations of DBSCAN and OPTICS~\footnote{\url{https://scikit-learn.org/stable/modules/clustering.html}}. 
The released JAVA implementation of random projection-based DBSCAN~\cite{rpDbscan} cannot run even on the small Mnist data set.
To demonstrate the scalability of sDBSCAN on other distance measures, we compare it with the popular scikit-learn $k$-means++~\cite{KmeanPlus} and the result of kernel $k$-means in~\cite{KernelKmean_Nys}.

We use the normalized mutual information (NMI) to measure the clustering quality.
We conduct experiments on three popular data sets: Mnist ($n = 70,000, d = 784$, \# clusters = 10), Pamap2 ($n = 1,770,131, d = 51$, \# clusters = 18), and Mnist8m ($n = 8,100,000, d = 784$, \# clusters = 10). 
We use the class labels as the ground truth.
For Pamaps2, we discarded instances that contain NaN values and removed the dominated class~0  corresponding to the transient activities.
We note that Mnist and Mnist8m are sparse data sets with at least 75\% sparsity, while Pamap2 is dense.
All results are the average of 5 runs of the algorithms.


\textbf{A note on our implementation.} We use the Eigen library~\footnote{https://eigen.tuxfamily.org} for SIMD vectorization on computing the distances.
Our sDBSCAN and sOPTICS are multi-threading friendly.
We only add \textbf{\texttt{\#pragma omp parallel}} directive on the \textbf{\texttt{for}} loop when preprocessing and finding the neighborhood for each point.

\subsection{An ablation study of sOPTICS on Mnist}

\begin{figure*} [t]
	\centering
	\includegraphics[width=1\textwidth]{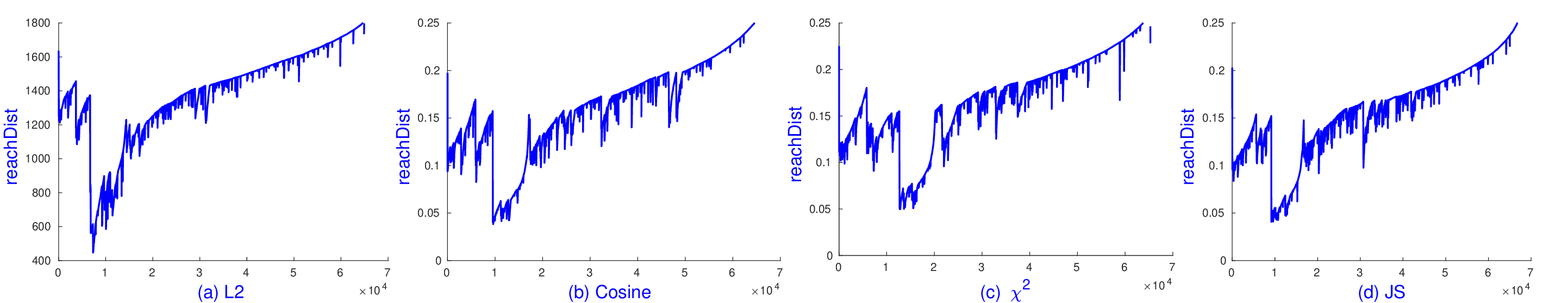}
	\caption{sOPTICS's graphs on L2, cosine, $\chi^2$, and JS on Mnist. Each runs in less than \textit{3 seconds}.}
	\label{fig:sOptics-Mnist}
\end{figure*} 

This subsection measures the performance of sOPTICS and scikit-learn OPTICS with multi-threads.
We show that sOPTICS can output the same OPTICS's results in less than \textit{30 seconds}, which is up to $180\times$ speedups over scikit-learn OPTICS.

\textbf{Comparison to scikit-learn OPTICS on L2 and L1.}
We use scikit-learn OPTICS to generate the dendrogram given $minPts = 50$.
To select a reasonable value of $\eps$ for OPTICS to reduce the running time, we randomly sample 100 points and use the average top-100 nearest neighbor distances of these sampled points as $\eps$.
Accordingly, we use $\eps = 1,800$ for L2 and $\eps = 18,000$ for L1.

To recover OPTICS's results, sOPTICS uses $D = 2,048, k = 2$ and $m = 1,000$. 
For L2 and L1 distances, sOPTICS requires two additional parameters, including the scale $\sigma$ and the number of embeddings $d'$, for random Fourier embeddings.
We observe that the performance of sOPTICS is insensitive to $\sigma$ and $d'$, hence we simply set $\sigma = 2 * \eps$ and $d' = 1,024$ for our experiments.

We run both scikit-learn OPTICS and sOPTICS with 64 threads.
Regarding speed, sOPTICS needs less than 30 seconds, while OPTICS takes hours to finish.
This shows the advantages of sOPTICS in accurately finding candidates to construct neighborhoods and efficiently utilizing multi-threading.
Regarding accuracy, sOPTICS outputs nearly the same graphs as OPTICS, as shown in Figure~\ref{fig:Optic-Mnist}.
Since there are more valleys on L2 than L1, the clustering accuracy regarding the ground truth using L2 will be higher than L1.

\textbf{On the needs of sOPTICS on other distances.}
While DBSCAN on L2 is popular, the capacity to use arbitrary distances is an advantage of DBSCAN compared to other clustering algorithms.
We use sOPTIC to visualize the cluster structure on other distance measures.
We use $D = 1,024, m = minPts = 50, k = 5$, and set
$\eps = 1,800$ for L2, and $\eps = 0.25$ for the others.
We use $d' = 1,024$ random features and set $\sigma = 2 * \eps$ for L2.

Figure~\ref{fig:sOptics-Mnist} shows reachability-plot dendrograms of sOPTICS on 4 studied distance measures.
Since points belonging to a cluster have a small reachability distance to their nearest neighbors, the number of valleys in the dendrograms reflects the cluster structure. 
From Figure~\ref{fig:sOptics-Mnist}, we can predict that cosine, $\chi^2$ and JS provide higher clustering quality than L2.
sOPTICS on these measures show a clear 4 valleys compared to 3 hazy valleys of L2.
Importantly, any $\eps$ in the range $[0.1, 0.15]$ can differentiate these 4 valleys while it is impossible to select a specific value to separate the 3 valleys of L2.  

Note that the sOPTICS graphs on L2 of Figures~\ref{fig:Optic-Mnist}(b) and~\ref{fig:sOptics-Mnist}(a) are slightly different due to the smaller values of $D, m$.
Hence, sOPTICS in the setting in Figure~\ref{fig:sOptics-Mnist} runs significantly faster.
As sOPTICS graph on L1 shows an inferior clustering quality, we do not report L1.

\subsection{An ablation study of sDBSCAN on Mnist}

This subsection measures the performance of sDBSCAN compared to scikit-learn DBSCAN with multi-threads.
We also compare sDBSCAN with recent competitors, including pDBSCAN~\cite{DBSCANVisit1}, DBSCAN++~\cite{DBSCAN++} (uDBSCAN++ and kDBSCAN++), and sngDBSCAN~\cite{sngDBSCAN} with 1 thread as their implementations do not support multi-threading.

\subsubsection{Comparison to DBSCAN's output on cosine distance.}
Given $minPts = 50$, we set $\eps = 0.11$ as DBSCAN returns the highest NMI of 43\% compared to the ground truth.
sDBSCAN has two main parameters on cosine distance: top-$m$ points closest/furthest to a random vector and top-$k$ closest/furthest random vectors to a point.
These parameters govern the accuracy and efficiency of sDBSCAN.
We set $D = 1,024$ and vary $m, k$ in the next experiments.

\begin{table}[!b]
    \centering
    \caption{Comparison of sDBSCAN with the DBSCAN's output on cosine distance with $\eps = 0.11, minPts = 50$ over different $k$ and $m$ on Mnist. The scikit-learn DBSCAN runs in \textit{71 seconds}.}
    \label{tab:sDBSCAN-DBSCAN}
    \begin{tabular}{c c c c c c c c}
    \toprule
         $m$ & 50 & 100 & 200 & 400 & 1,000 & 2,000 & \textbf{2,000}\\
         $k$ & 40 & 20 & 10 & 5 & 2 & 1 & \textbf{2}\\
         \midrule
         NMI & 69\%  & 76\% & 81\% & 86\% & 88\% & 91\% & \textbf{95\%}\\
         Time (s) & 8.8 & 9.9 & 10.5 & 11.1 & 12.1 & 12.6 & 23 \\
         \bottomrule
    \end{tabular}    
\end{table}

\textbf{Accuracy and efficiency of sDBSCAN.} 
For sDBSCAN, we vary $k = \{40, 20, \ldots, 1 \}$, $m = \{50, 100, \ldots, 2,000 \}$ such that it computes nearly the same $2km$ distances for each point.
Table~\ref{tab:sDBSCAN-DBSCAN} shows the performance of multi-threading sDBSCAN on DBSCAN's outputs with $\eps = 0.11, minPts = 50$.
We can see that sDBSCAN can recover DBSCAN's result with larger $m$.
These findings justify our theoretical result as
larger $m$ increases the chance of examining all points in the set $S_i$ and $R_i$ corresponding to the random vector $\br_i$, increasing the chance to recover the DBSCAN's output.

In the last column with $m = 2,000$, sDBSCAN with $k=2$ achieves NMI of 95\% but runs 2 times slower than $k=1$ since it nearly doubles the candidate size. 
sDBSCAN with all configurations runs at least $3\times$ faster than scikit-learn DBSCAN. 
This presents the advantages of sDBSCAN in accurately finding candidates to find core points and efficiently utilizing the multi-threading architecture.


\textbf{sDBSCAN's parameter setting.}
Given $m \geq minPts$ and $k$, each point needs at most $2km$ distance computations due to the duplicates on top-$m$ candidates corresponding to $2k$ investigated random vectors.
As a larger $k$ leads to more duplicates among $2km$ candidates, given a fixed budget $B = 2km$, Table~\ref{tab:sDBSCAN-DBSCAN} shows that $k = 1, m = B/2$ has higher accuracy but significantly higher running time than $k = k_0 > 1, m = B/2 k_0$.

\begin{table}[b]
    \centering
        \caption{Running time of sDBSCAN components in seconds with $D = 1,024, k = 5, m = minPts = 50, \eps = 0.11$ on Mnist.}
    \label{tab:sDBSCAN-Time}
    \begin{tabular}{c c c c c}
        \toprule
        \# Threads & Preprocess & Find core points & Cluster & Total \\
        \midrule
        1 thread & 1.424 & 7.745 & 0.001 & 9.198 \\
        64 threads & 0.160 & 0.700 & 0.001 & 0.862 \\
        \bottomrule
    \end{tabular}
\end{table}

While larger $m$ and $D$ tend to increase the accuracy of sDBSCAN, they affect space and time complexity.
Since we often set $D = 1,024$ to ensure Lemma~\ref{lm:core} holds,
and the memory resource is limited due to large data sets, we set $m = minPts = 50$ and $k = \{5, 10\}$ for most experiments.
Compared to the ground truth, this setting does not affect the accuracy of sDBSCAN but uses significantly less computational resources than other configurations.

\begin{figure*} [ht]
	\centering
	\includegraphics[width=1\textwidth]{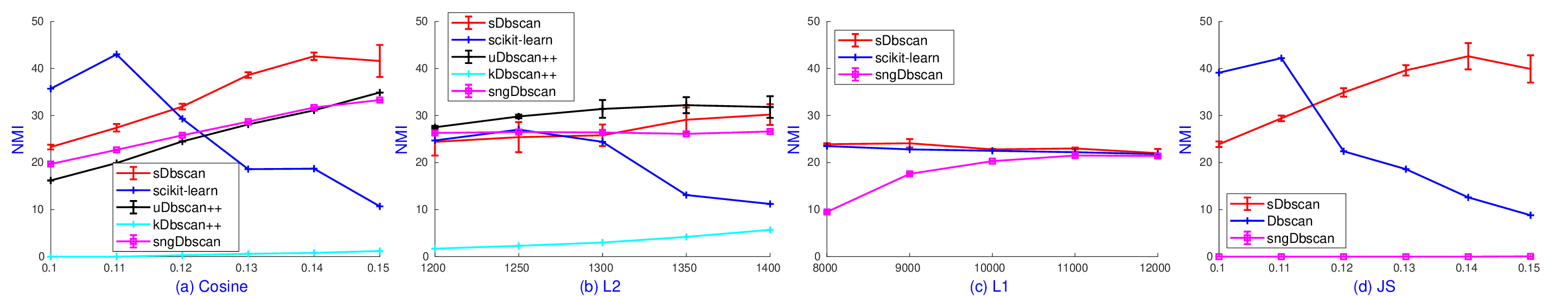}
	\caption{NMI comparison of DBSCAN variants on cosine, L2, L1, JS distances on Mnist over suggested ranges of $\eps$ by sOPTICS in Figure~\ref{fig:sOptics-Mnist}. pDBSCAN and scikit-learn have identical results. As the sampling-based DBSCAN implementations only support L2 and cosine distances, we report our implemented sngDBSCAN on L1 and JS.}
	\label{fig:mnist}
\end{figure*}

\begin{table*}[!ht]
    \centering
    \caption{The NMI on the best $\eps \in [0.1, 0.2]$ and running time comparison of DBSCAN variants using cosine distance with $minPts = 50$ on Mnist. sDBSCAN and scikit-learn DBSCAN run in multi-threading while the others are on a single thread.}
    \label{tab:Mnist}
     \begin{tabular}{c c c c c c c c}   
     \toprule
        Alg. & sDBSCAN & scikit DBSCAN & sDBSCAN$_{1}$ & uDBSCAN++ & kDBSCAN++ & sngDBSCAN & pDBSCAN \\
        \midrule
        NMI & \textbf{43\%} & \textbf{43\%} & \textbf{43\%} & \textbf{43\%} & 7\% & 33\% & \textbf{43\%} \\
        Time & \textbf{0.9s} & 86s & \textbf{8.8s} & 67s & 87s & 18s & $1.85$ hours \\ 
        $\eps$ & 0.14 & 0.11 & 0.14 & 0.18 & 0.2 & 0.15 & 0.11 \\
        \bottomrule
    \end{tabular}
    
\end{table*}

\textbf{Running time of sDBSCAN's components.}
Given $\eps=0.11$, $m = minPts = 50$, $k = 5$,  
Table~\ref{tab:sDBSCAN-Time} shows the running time of sDBSCAN components on 1 and 64 threads.
On 1 thread, we can see that finding neighborhoods is the primary computational bottleneck while forming clusters is negligible.
This is due to the expensive random access operations to compute $2km$ distances for each point.
As the main components of sDBSCAN, including preprocessing (Alg.~\ref{alg:preprocessing}) and finding neighborhoods (Alg.~\ref{alg:neighborhood}) can be sped up with multi-threading, we achieve nearly $10\times$ speedup with 64 threads.
We also observe that the construction time of the randomized embeddings to support L2, L1, $\chi^2$, and JS is similar to the preprocessing time and is negligible to that of finding neighborhoods.

\subsubsection{Comparison to the ground truth on several distances.}
We conduct experiments to measure the clustering accuracy regarding the class labels as the ground truth.
We fix $minPts = 50$, and select a suitable range values of $\eps$ based on the sOPTICS's graphs.
For example, we select 5 different values $\eps$ on $[0.1, 0.15]$ for $\chi^2$ and JS, and report the representative result of JS due to the similar outcomes.

sDBSCAN uses $D = 1,024$, $m = 50$, $k = 5$, and hence computes at most $2km = 1000$ distances for each point.
For DBSCAN++ variants and sngDBSCAN, we set the sampling probability $p = 0.01$ so that they compute $pn = 700$ distances for each point.
pDBSCAN uses $\rho = 0.001$ as we observe that changing $\rho = 0.0001$ does not affect its running time.
We also found that the brute force for range search queries is the fastest method for exact DBSCAN on Mnist.

Figure~\ref{fig:mnist} shows NMI scores over a wide range of $\eps$ on 4 different distances.
The grid-based approach pDBSCAN only supports L2 and its accuracy is identical to scikit-learn on cosine and L2.
sDbscan is superior on all 4 supported distances, except L2 when compared to uDBSCAN++.
While sDBSCAN reaches DBSCAN's accuracy of NMI 43\% on cosine and JS, sngDBSCAN gives lower accuracy on all 4 distances.
While uDBSCAN++ gives at most 32\% NMI on L2 and cosine, kDBSCAN++ does not provide a meaningful result on the studied range values of $\eps$.
L2 and L1 distances show inferior performance on clustering compared to cosine and JS, as can be predicted from their sOPTICS graphs in Figure~\ref{fig:sOptics-Mnist}.

\textbf{Comparison to the ground truth on cosine distance over a wider range of $\eps$.}
Table~\ref{tab:Mnist} summarizes the performance of studied DBSCAN variants on the best $\eps \in [0.1, 0.2]$ with cosine distance.
sDBSCAN has two variants: multi-threading and single thread (sDBSCAN$_1$).
On 1 thread, sDBSCAN$_1$ runs nearly $2\times, 8\times$ and $10\times$ faster than sngDBSCAN, uDBSCAN++ and kDBSCAN++, respectively.
This is due to the significant overhead of the KD-Tree indexes used by uDBSCAN++ and $k$-center initialization of kDBSCAN++.
Though pDBSCAN shares the same NMI with scikit-learn DBSCAN, it is not feasible for high-dimensional data sets.
It takes nearly 2 hours to finish, while the others need less than 1.5 minutes.

sDBSCAN achieves the same NMI of 43\% with DBSCAN.
Its multi-threading sDBSCAN runs in less than \textit{1 second}, which is $96\times$ faster than multi-threading scikit-learn.
While sDBSCAN uses the range of $\eps$ suggested by sOPTICS, sampling-based approaches have to explore a much wider range for $\eps$ to recover DBSCAN's accuracy.

\begin{figure*} [t]
	\centering
	\includegraphics[width=1\textwidth]{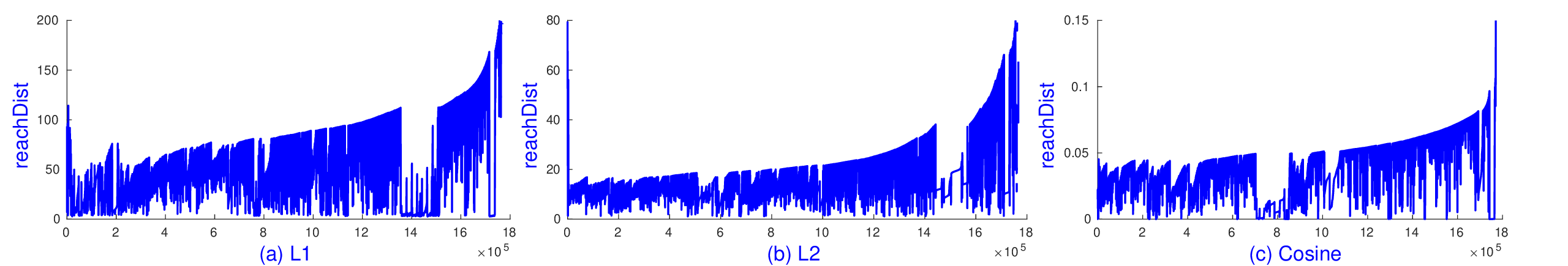}
	\caption{sOPTICS's graphs on L1, L2 and cosine distances on Pamap2. Each runs in less than \textit{2 minutes}.}
	\label{fig:sOptics-Pamap}
\end{figure*} 
\begin{figure*} [t]
	\centering
	\includegraphics[width=1\textwidth]{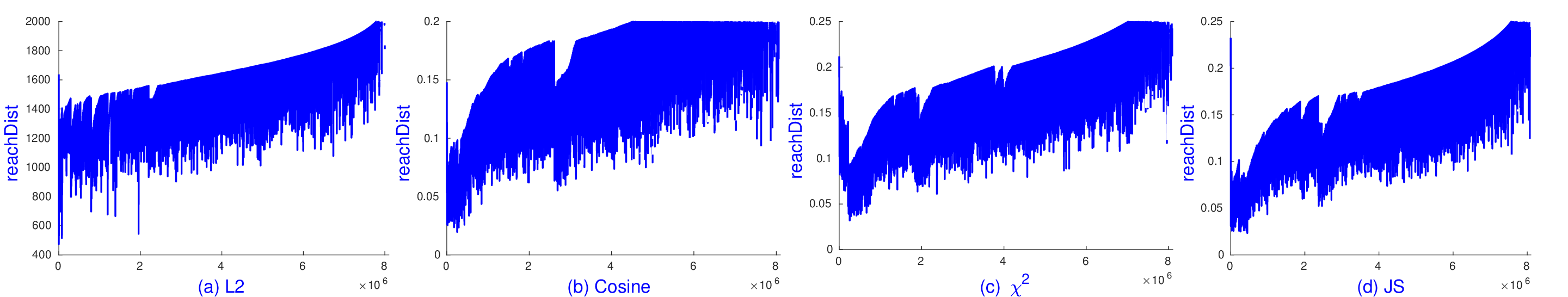}
	\caption{sOPTICS's graphs on L2, cosine, $\chi^2$, and JS distances on Mnist8m. Each runs in less than \textit{11 minutes}.}
	\label{fig:sOptics-Mnist8m}
\end{figure*}

\subsection{Comparison on Pamap2 and Mnist8m}

This subsection compares the performance of sDBSCAN and sOPTICS with other clustering variants on million-point Pamap2 and Mnist8m data sets.
scikit-learn OPTICS and DBSCAN cannot finish due to the memory constraint, and the grid-based pDBSCAN cannot finish after 2 hours.
As our implemented sngDBSCAN runs significantly faster than~\cite{sngDBSCAN} and supports multi-threading, we use our implementation for comparison.
We set $p = 1$ for sngDBSCAN to recover the exact DBSCAN.
sDBSCAN and sOPTICS use $k = 10, m = minPts = 50$.
DBSCAN++ variants and sngDBSCAN use $p = 0.01$ on Pamap2 and $p = 0.001$ on Mnist8m.


\subsubsection{sOPTICS graphs on Pamap2 and Mnist8m.}

\begin{figure*} [!t]
	\centering
	\includegraphics[width=1\textwidth]{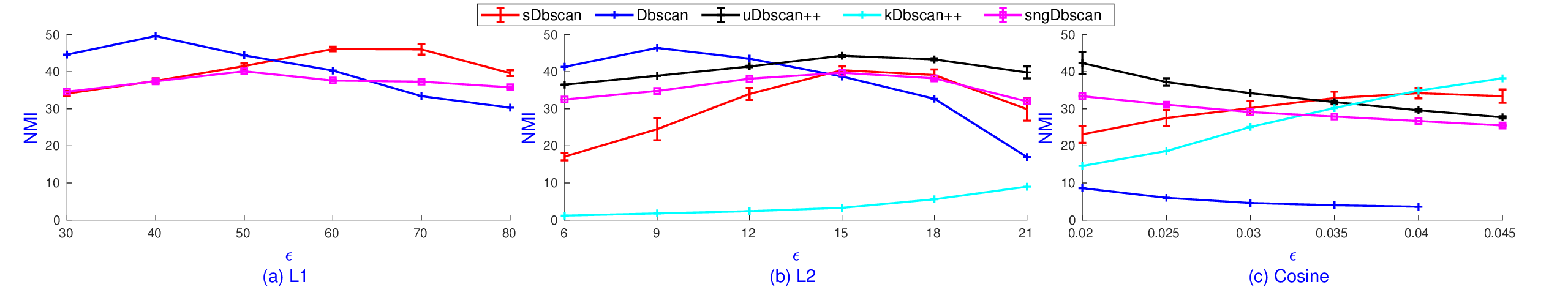}
	\caption{NMI comparison of DBSCAN variants on L1, L2 and cosine on Pamap2 on suggested ranges of $\eps$ by
sOPTICS in Figure~\ref{fig:sOptics-Pamap}.}
	\label{fig:sDbscan-Pamap}
\end{figure*}

We use the average top-100 nearest neighbor distances of 100 sampled points to select a suitable $\eps$ for sOPTICS.
Figure~\ref{fig:sOptics-Pamap} and~\ref{fig:sOptics-Mnist8m} show sOPTICS's graphs of Pamap2 and Mnist8m with $minPts = 50$.
 As Pamap2 contains negative features, we only run on cosine, L2, and L1 distances.
sOPTICS uses $D = 1,024$, $\sigma = 2 * \eps$, $d' = 1,024$ runs in less than \textit{2 minutes} and \textit{11 minutes} on Pamap2 and Mnist8m, respectively.
Both figures show that L2 is less relevant than the other distances as it does not show clear valley areas.
This reflects the need to use other distance measures rather than L2 for density-based clustering to achieve reasonable performance.

\subsubsection{Comparison to the ground truth on Pamap2.} 
From the sOPTICS graphs, we select relevant range values for $\eps$ to run DBSCAN variants.
Figure~\ref{fig:sDbscan-Pamap} shows the NMI scores of studied algorithms on such ranges.
On these ranges of $\eps$, the NMI peak of sDBSCAN is consistently higher than that of sngDBSCAN but is 5\% lower than DBSCAN on L1 and L2.
While sDBSCAN achieves the peak of NMI at the investigated ranges of $\eps$ on 3 distances, the performance of the others is very different on cosine.
We found that DBSCAN with $\eps = 0.005$ returns NMI 47\% but offers only 37\% NMI on $\eps = 0.01$.
This explains the reliability of sOPTICS on guiding the parameter setting for sDBSCAN and the difficulty in selecting relevant $\eps$ for other DBSCAN variants without any visual tools.

\begin{table}[hb]
	\centering	
      \caption{The NMI on the best $\eps$ and running time comparison on cosine and L1 distances on Pamap2. The upper 3 algorithms run in multi-threading with $10\times$ speedup compared to 1 thread while the lower ones use 1 thread.}
    \label{tab:Pamap}
    
	\begin{tabular}{c  c c  c  c c c} 
		\toprule
		\multirow{2}{*}{Algorithms} & \multicolumn{3}{c}{Cosine ($\eps \in [0.005, 0.05]$)} & \multicolumn{3}{c}{L1 ($\eps \in [30, 80]$)} \\ 
		\cmidrule{2-7} 
		& NMI & Time & $\eps$ & NMI & Time & $\eps$ \\ 
		\midrule
        DBSCAN & \textbf{47\%}  & 28.4 min & 0.005 & \textbf{50\%} & 29.3 min & 40\\
        sDBSCAN & 34\% & \textbf{0.2 min} & 0.04 & \textbf{46\%} & \textbf{0.3 min} & 60 \\
        sngDBSCAN & 42\% & 2.8 min & 0.015 & 40\%  & 2.7 min & 50\\ 
		\midrule
		uDBSCAN++ & \textbf{46\%} & 3 min & 0.015 & --  & -- & --\\ 
        kDBSCAN++ & 39\% & 13.4 min & 0.05 & --  & -- & --\\ 
        $k$-mean++ ($k = 18$) & 36\%  & 0.4 min & -- & -- & -- & -- \\
		 \bottomrule
	\end{tabular}

\end{table}

\begin{figure*} [!t]
	\centering
	\includegraphics[width=1\textwidth]{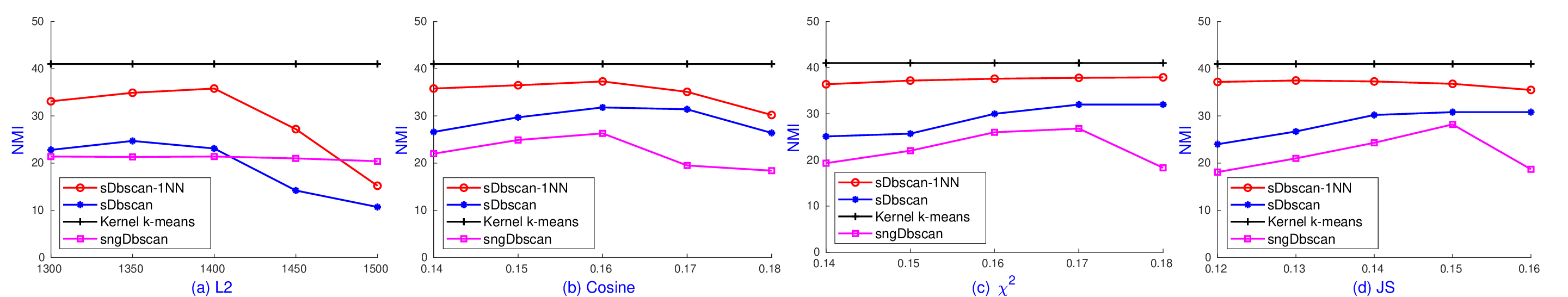}
	\caption{NMI comparison of DBSCAN variants on L2, cosine, $\chi^2$, and JS distances and kernel $k$-means on Mnist8m over suggested ranges of $\eps$ by
sOPTICS in Figure~\ref{fig:sOptics-Mnist8m}.}
	\label{fig:sDbscan-Mnist8m}
\end{figure*} 
\begin{table*}[!t]
	\centering
      \caption{The NMI on the best $\eps$ and running time comparison of multi-threading DBSCAN variants on L2, cosine, $\chi^2$, and JS distances on Mnist8m. Kernel $k$-means ($k = 10$)~\cite{KernelKmean_Nys} runs in 15 minutes on a supercomputer of 32 nodes and achieves NMI 41\%.}
    \label{tab:Mnist8m}
	\begin{tabular}{c  c c  c  c c c  c c c c c c} 
		\toprule
		\multirow{2}{*}{Algorithms} & \multicolumn{3}{c}{L2 ($\eps \in [1100, 1500]$)} & \multicolumn{3}{c}{Cosine ($\eps \in [0.1, 0.2]$)} & \multicolumn{3}{c}{$\chi^2$ ($\eps \in [0.1, 0.2]$)} & \multicolumn{3}{c}{JS ($\eps \in [0.1, 0.2]$)} \\ 
		\cmidrule{2-13} 
		& NMI & Time & $\eps$ & NMI & Time & $\eps$ & NMI & Time & $\eps$ & NMI & Time & $\eps$ \\ 
		\midrule
		sDBSCAN-1NN & \textbf{36\%} & 8 min & 1400 & \textbf{37\%} & 14 min & 0.16 & \textbf{38\%} & 15 min & 0.17 & \textbf{38\%} & 15 min & 0.15 \\ 
		sDBSCAN & 25\% & \textbf{7 min} & 1350 & 32\% & \textbf{8 min} & 0.16 & 32\% & \textbf{10 min} & 0.17 & 31\% & \textbf{10 min} & 0.15 \\ 
        sngDBSCAN ($p = 0.001$) & 22\% & 43 min & 1150 & 26\% & 42 min & 0.16 & 28\%  & 64 min & 0.15 & 27\%  & 64 min & 0.17 \\        
		 \bottomrule
	\end{tabular}
\end{table*}

As L2 is inferior to L1 and cosine on sampling-based DBSCAN, Table~\ref{tab:Pamap} reports the NMI scores and running time of studied algorithms on L1 and cosine distances over a larger range of $\eps$. 
While sDBSCAN shows a low NMI on cosine, its accuracy is 45\% on L1, which is higher than sngDBSCAN on both cosine and L1.
As sngDBSCAN uses $p = 0.01$, each point will compute $np \sim 17,700$ distances compared to $2km = 1000$ of sDBSCAN.
Hence, sDBSCAN runs up to 14$\times$ faster than sngDBSCAN.
On L1, sDBSCAN is nearly $100\times$ faster than DBSCAN.
By changing $k = 5, m = 200$, sDBSCAN reaches 48\% NMI on L1 and still runs in less than \textit{1 minute}.
Compared to $k$-mean++, multi-threading sDBSCAN runs faster and offers significantly higher NMI with L1.
We emphasize that the advantage of multi-threading sDBSCAN comes from its simplicity, as it is effortless engineering to run sDBSCAN on multi-threads.

Among sampling-based approaches, uDBSCAN++ gives the highest NMI score of 46\%, while kDBSCAN++ is very slow.
kDBSCAN++ runs approximately $5\times$ slower than uDBSCAN due to the overhead of $k$-center initialization and the efficiency of KD-Trees on Pamap2 with $d = 51$.
In general, sDBSCAN provides competitive clustering accuracy and runs significantly faster than other DBSCAN variants.
Like Mnist, our sOPTICS on Pamap2 runs in less than \textit{2 minutes}, even faster than any sampling-based implementations.

\subsubsection{Comparison to the ground truth on Mnist8m.}
As Mnist8m is non-negative, we conduct experiments on cosine, L2, and L1, $\chi^2$, and JS distances.
Similar to Mnist, sOPTICS graphs show L1 is inferior, so we do not report L1 here.

uDBSCAN++ and kDBSCAN++ with $p = 0.001$ could not finish after 3 hours.
It is not surprising as our multi-threading sngDBSCAN runs in nearly 1 hour, and single thread sngDBSCAN is at least $4\times$ faster than DBSCAN++ variants on Mnist as shown in Table~\ref{tab:Mnist}.
As we cannot run scikit-learn $k$-means++, we use the result of 41\% NMI given by a fast kernel $k$-means~\cite{KernelKmean_Nys} running on a supercomputer with 32 nodes, each of which has two 2.3GHz 16-core (32 threads) Haskell processors and 128GB of DRAM.
This configuration of a \textit{single} node is similar to our conducted machine.


From the sOPTICS graphs, we select 5 specific values on the relevant range of $\eps$, including $\{1200, \ldots, 1400 \}$ for L2, $\{0.14, \ldots, 0.18\}$ for cosine, $\chi^2$, and $\{0.12, \ldots, 0.16\}$ for JS.
Figure~\ref{fig:sDbscan-Mnist8m} shows the NMI scores of 3 algorithms, sDBSCAN-1NN with 1-NN heuristic described in Subsection 4.5, sDBSCAN, sgnDBSCAN, and kernel $k$-means on Mnist8m.
sDBSCAN-1NN shows superiority among DBSCAN variants, achieving 10\% and 5\% higher than sngDBSCAN and sDBSCAN, respectively, at the peak of NMI on 4 studied measures.
sDBSCAN consistently gives higher accuracy than sngDBSCAN with the most significant gap of 5\% on $\chi^2$ and cosine.

Table~\ref{tab:Mnist8m} summarizes the performance of multi-threading sDBSCAN-1NN, sDBSCAN, and sngDBSCAN, including the NMI score on the best $\eps$ and the running time, on L2, cosine, $\chi^2$, and JS.
sDBSCAN-1NN runs in at most 15 minutes and returns the highest NMI among DBSCAN variants with a peak of 38\% NMI on $\chi^2$ and JS.
We emphasize that kernel $k$-means with Nystr\"{o}m approximation~\cite[Table 4]{KernelKmean_Nys} also needs 15 minutes on a supercomputer and gives 41\% NMI.

sDBSCAN runs faster than sDBSCAN-1NN as it does not assign labels to non-core points, and still achieves significantly higher NMI scores than sngDBSCAN on all 4 studied distances.
Among 4 distances, L2 shows lower accuracy but runs faster than $\chi^2$ and JS due to the faster distance computation.
sDBSCAN runs $6.4\times$ faster than sngDBSCAN, which is justified by the number of distance computations $np = 8,100$ of sngDBSCAN compared to $2km = 1000$ of sDBSCAN.
As sDBSCAN-1NN samples $0.01n$ core points to build the approximate 1NN classifier, the running time overhead of this extra step is smaller than sDBSCAN's time.

\section{Conclusion}

The paper presents sDBSCAN for density-based clustering, and sOPTICS for interactive exploration of density-based clustering structure for high-dimensional data.
By leveraging the neighborhood preservation of random projections, sDBSCAN preserves the DBSCAN's output with theoretical guarantees.
We also extend our proposed algorithms to other distance measures to facilitate density-based clustering on many applications.
Empirically, both sDBSCAN and sOPTICS are very scalable, run in minutes on million-point data sets, and provide very competitive accuracy compared to other clustering algorithms.
We hope that the results of our work will stimulate research works that leverage lightweight indexing for ANNS to scale up density-based clustering in high dimensions.

\bibliographystyle{ACM-Reference-Format}
\bibliography{sigproc}

\appendix
\section{Random kernel features}

We first show empirical results on random kernel mappings that facilitate sDBSCAN and sOPTICS on L1, L2, $\chi^2$, and JS distance.
We use Pamap2 for L1 and L2 as it contains negative values and Mnist for $\chi^2$ and JS as it does not contain negative values.

\subsection{L1 and L2 on Pamap2}

\begin{figure} [h]
	\centering
	\includegraphics[width=1\columnwidth]{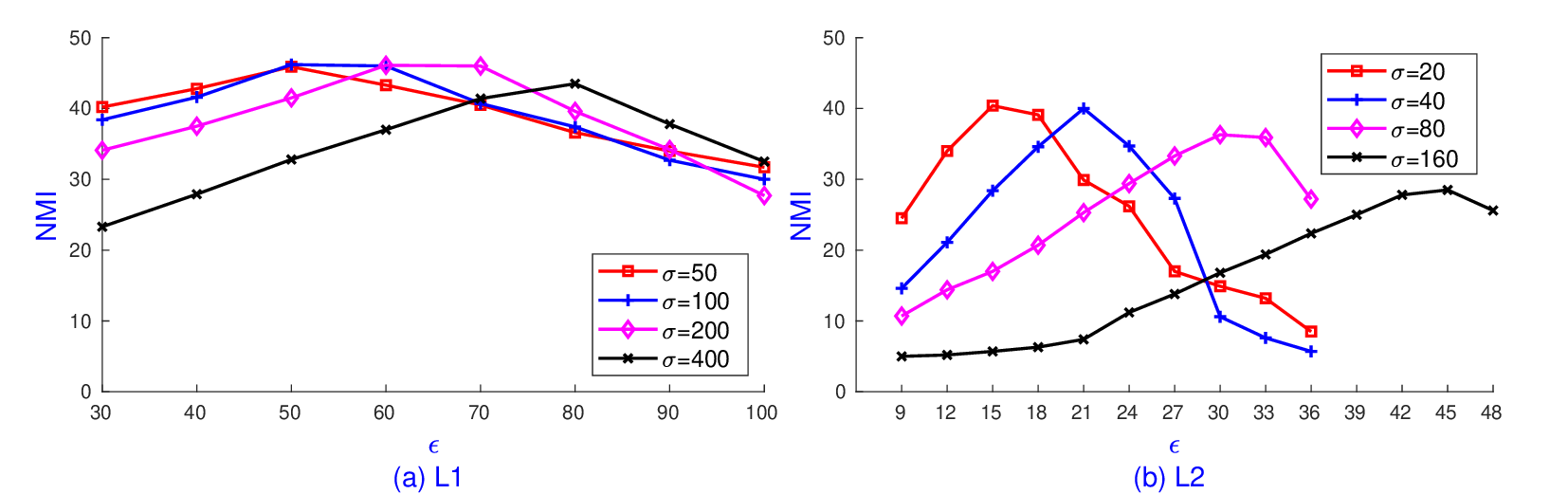}
	\caption{sDBSCAN's NMI on L1 and L2 with various $\sigma$ on Pamaps with $k = 10, m = 50$. Each runs in less than \textit{20 seconds}.}
	\label{fig:sDbscan-Pamap-sigma}
\end{figure}

\begin{figure*} [h]
	\centering
	\includegraphics[width=1\textwidth]{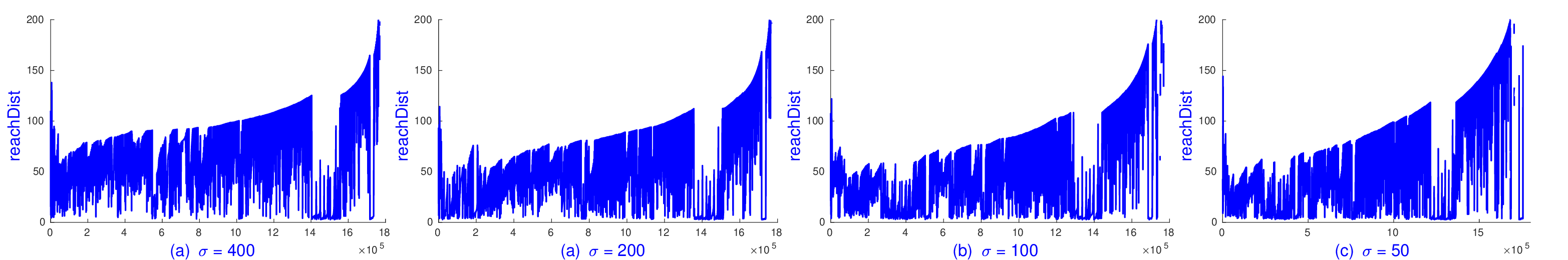}
	\caption{sOPTICS's graphs on L1 with various $\sigma$ on Pamaps with $k = 10, m = 50$. Each runs in less than \textit{2 minutes}.}
	\label{fig:sOptics-Pamap-L1}
\end{figure*}

\begin{figure*} [h]
	\centering
	\includegraphics[width=1\textwidth]{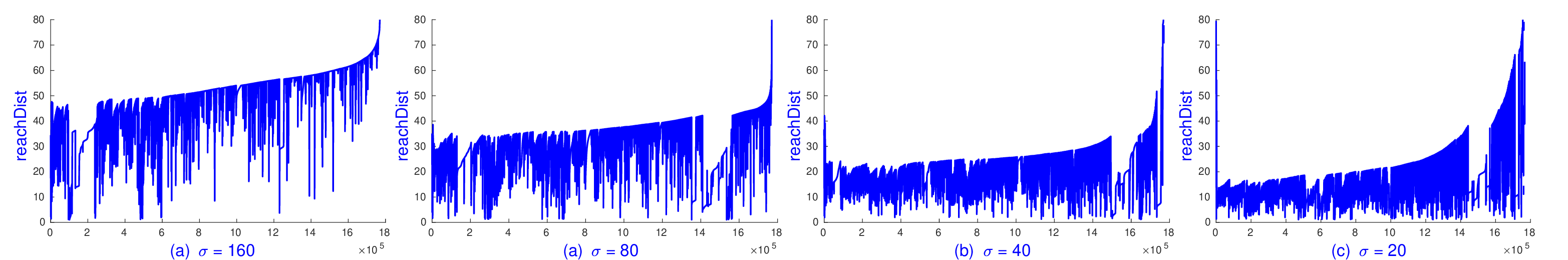}
	\caption{sOPTICS's graphs on L2 with various $\sigma$ on Pamaps with $k = 10, m = 50$. Each runs in less than \textit{2 minutes}.}
	\label{fig:sOptics-Pamap-L2}
\end{figure*}

We carry out experiments to evaluate the sensitivity of the parameter $\sigma$ used on random kernel mappings on L1 and L2.
We fix $k = 10, m = minPts = 50, D = 1024, d' = 1024$ and vary $\sigma$ for L1 and L2.
Figure~\ref{fig:sDbscan-Pamap-sigma} shows the accuracy of sDBSCAN on the recommended range of $\eps$ by sOPTICS graphs with $\sigma = \{50, 100, 200, 400$ for L1 on Figure~\ref{fig:sOptics-Pamap-L1} and $\sigma = \{20, 40, 80, 160\}$ for L2 on Figure~\ref{fig:sOptics-Pamap-L2}.

\begin{figure} [h]
	\centering
	\includegraphics[width=1\columnwidth]{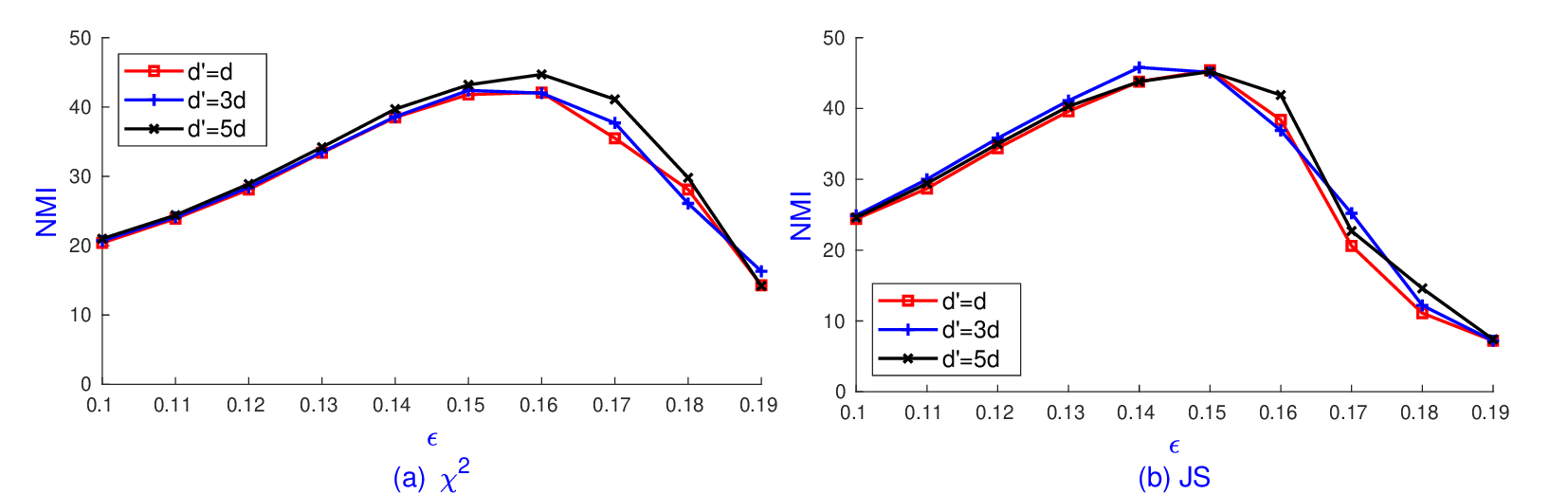}
	\caption{sDBSCAN's NMI on L1 and L2 with various $\sigma$ on Pamaps with $k = 10, m = 50$. Each runs in less than \textit{20 seconds}.}
	\label{fig:sDbscan-mnist-L3-L4}
\end{figure}

\begin{figure} [h]
	\centering
	\includegraphics[width=1\columnwidth]{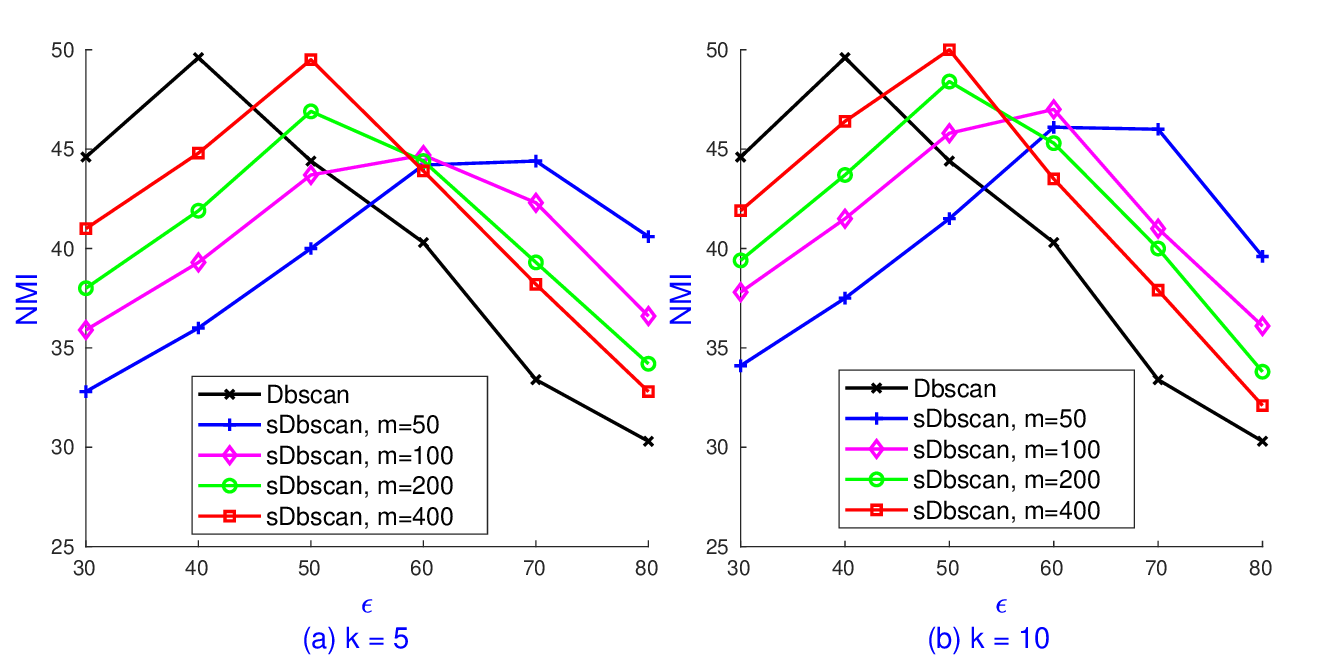}
	\caption{sDBSCAN's NMI on L1 with various $k, m$ on Pamaps.}
	\label{fig:sDbscan-L1-pamap0-k-m}
\end{figure}

While L1 provides 46\% NMI, higher than just 40\% NMI by L2, the setting $\sigma = 2 \eps$ reaches the peak of NMI for both L1 and L2.
For example, sDBSCAN at $\eps = 50, \sigma = 100$ reaches NMI 46\% for L1.
sDBSCAN at $\eps = 15, \sigma = 20$ and $\eps = 21, \sigma = 40$ reach NMI of 40\% for L2.
Also, the performance of sDBSCAN is not sensitive to the value $\sigma$, especially with the guide of sOPTICS graphs.
For example on L2 as shown in Figure~\ref{fig:sOptics-Pamap-L2}, $\sigma = 160$ suggests the suitable range $[40, 50]$ while $\sigma = 80$ shows $[30, 40]$. 
The best values of $\eps$ of these two $\sigma$ values are clearly on these ranges, as shown in Figure~\ref{fig:sDbscan-Pamap-sigma}(b).
This observations appear again on $\sigma = \{20, 40\}$ for the range of $[10, 20]$ and $[20, 30]$, respectively.

\begin{figure*} [h]
	\centering
 \includegraphics[width=1\textwidth]{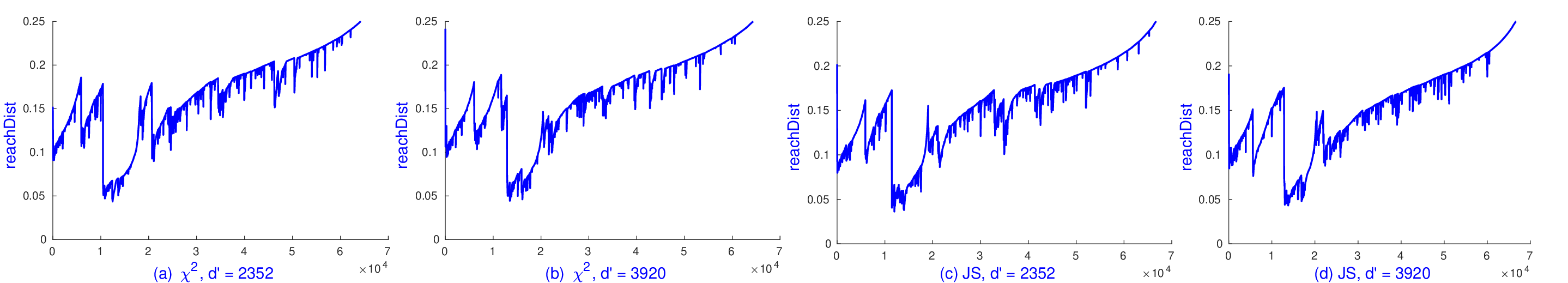}
	\caption{sOPTICS's graphs on $\chi^2$ and JS with $d' = \{3d, 5d\}$ on Mnist with $k = 5, m = 50$. Each runs in less than \textit{3 seconds}.}
	\label{fig:sOptics-Mnist-L3-L4}
\end{figure*} 

\subsection{$\chi^2$ and JS on Mnist}

We carry out experiments to evaluate the sensitivity of the parameter $d'$ used on random kernel mappings on $\chi^2$ and JS.
We fix $k = 5, m = minPts = 50, D = 1024$ and vary $d'$.
We note that ~\cite{PAMI12} approximates $\chi^2$ and JS distance using the sampling approach, hence we set the sampling interval as 0.4, as suggested on scikit-learn~\footnote{https://github.com/scikit-learn/scikit-learn/blob/5c4aa5d0d/sklearn/kernel\_approximation.py}. 
We also note that $d'$ should be set as $(2l + 1)d$ for $l \in \mathbb{N}$.
Hence, we set $d' = \{3d, 5d \}$ in our experiment.

Figure~\ref{fig:sOptics-Mnist-L3-L4} shows sOPTICS graphs of $\chi^2$ and JS on $d' = \{3d, 5d \}$.
They are very similar to the sOPTICS graphs on $d' = d$ where the range of $\eps$ should be in $[0.15, 0.2]$ for $\chi^2$ and $[0.13, 0.18]$ for JS.
Therefore, we can see that sDBSCAN with such a recommended range of $\eps$ will reach the peak of accuracy.

Figure~\ref{fig:sDbscan-mnist-L3-L4} shows that sDBSCAN reaches the peak at $\eps = 0.16$ with $\chi^2$ and $\eps = 0.14$ with JS.
Using $d' = 5d$ slightly increases the accuracy compared to $d' = d$ though it is not significant on JS.
Both measures offer the highest accuracy with 45\% NMI.

\section{Sensitivities of \lowercase{\textit{k, m}}}

We carry out experiments on Pamap2 using L1 to study the performance of sDBSCAN on various values of $k, m$.
We fix $\sigma = 200, D = 1024$ and consider the range of $\eps \in [30, 80]$.
As increasing $k, m$ will increase the memory and running time of sDBSCAN, we first fix $k = \{5, 10\}$ and then increase $m = \{50, 100, 200, 400\}$.

Figure~\ref{fig:sDbscan-L1-pamap0-k-m} shows that for a fix $k$, increasing $m$ slightly increases the accuracy of sDBSCAN.
In particular, sDBSCAN with $m = 400$ at $\eps = 50$ reaches the accuracy of the exact DBSCAN of 50\% NMI.
Regarding the time, sDBSCAN with $m = 400$ needs \textit{1.5 minutes} and \textit{xxx mins} for $k = 5$ and $k = 10$, respectively, which is significantly faster than \textit{29.3 minutes} required by the exact DBSCAN.

\section{Neighborhood size \lowercase{\textit{min}}\textit{P}\lowercase{\textit{ts}} = 100}

We present experiments on the setting of $minPts = 100$. 
We follow the same procedure that plots sOPTICS graphs first and use them to select the relevant range values of $\eps$.
We all use $k = 10, m = 100, D = 1024$ for Pamap2 and Mnist8m.

\subsection{Pamap2}

\begin{figure*} [t]
	\centering
	\includegraphics[width=1\textwidth]{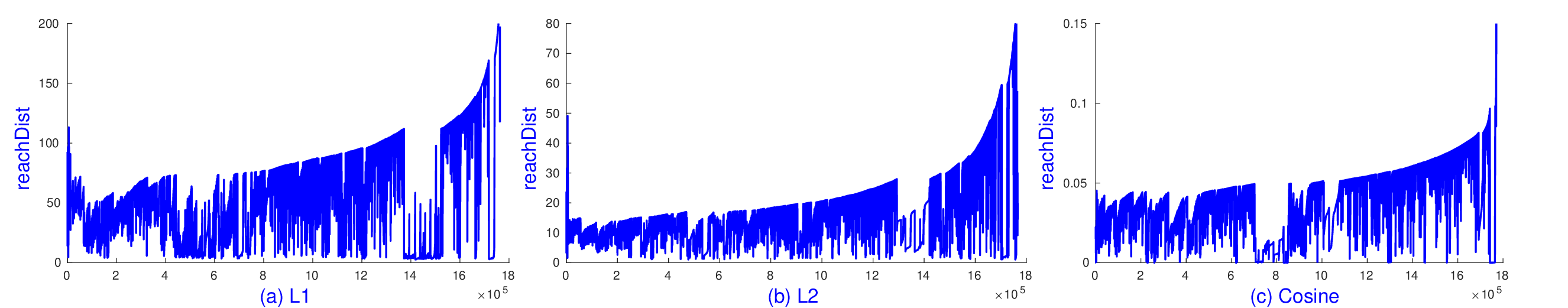}
	\caption{sOPTICS's graphs on L1, L2, cosine on Pamap2 with $k = 10, m = 100$. Each runs in less than \textit{5 minutes}.}
	\label{fig:sOptics-Pamap-100}
\end{figure*}

\begin{figure*} [h]
	\centering
 \includegraphics[width=1\textwidth]{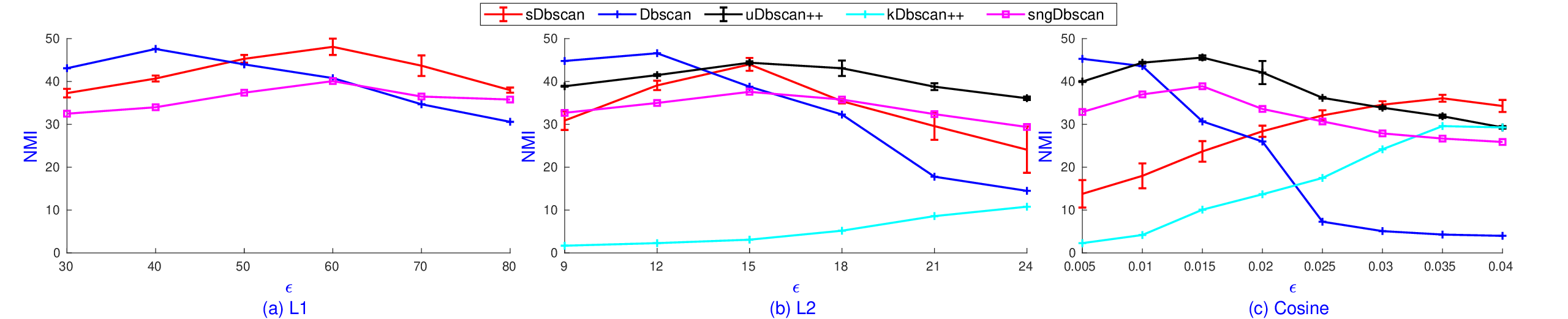}
	\caption{sDBSCAN's NMI on L1, L2, and cosine on Pamap2 with $k = 10, m = 100$.}
	\label{fig:sDbscan-Pamap-100}
\end{figure*}

Figure~\ref{fig:sOptics-Pamap-100} shows sOPTICS graphs on Pamap2 with L1, L2, and cosine distances where we use $\sigma = 200$ for L1, and $\sigma = 20$ for L2.
It shows again that L2 is inferior than L1 and cosine distances.
Figure~\ref{fig:sDbscan-Pamap-100} shows the accuracy of sDBSCAN compared to other competitors on Pamap2 with studied distances over the range of $\eps$ suggested by their corresponding sOPTICS graphs.

It is clear that sDBSCAN reaches the highest NMI at the suggested range of $\eps$ by their sOPTICS graphs.
In contrast, sampling-based approaches have to investigate a much wider range of $\eps$ to achieve good performance.
We observe that the performance of sDBSCAN is stable regarding $minPts$ though a larger $minPts$ requires larger $m$ which increases the running time.
Each instance of sDBSCAN runs in 0.5 minutes, which is significantly faster than 3 minutes by uDBSCAN++ and 13.4 minutes by kDBSCAN++.
sDBSCAN shows the advantage of running on many distance measures, which leads to the highest of 48\% of NMI of L1.
In contrast, uDBSCAN++ achieves the highest of 46\% of NMI among cosine and L2 distances.

\begin{figure*} [t]
	\centering
	\includegraphics[width=1\textwidth]{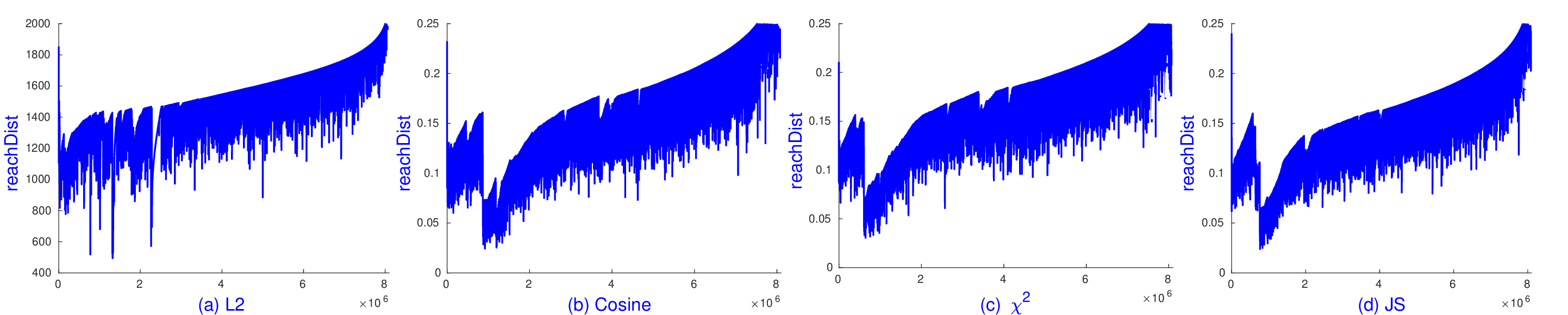}
	\caption{sOPTICS's graphs on L2, cosine, $\chi^2$, and JS distances on Mnist8m with $k = 10, m = 100$. Each runs in less than \textit{20 minutes}.}
	\label{fig:sOptics-Mnist8m-100}
\end{figure*} 

\begin{figure*} [t]
	\centering
 \includegraphics[width=1\textwidth]{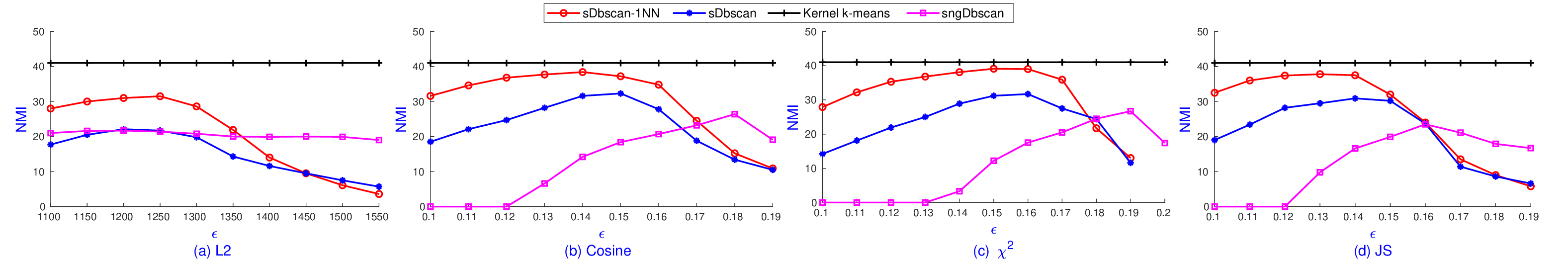}
	\caption{sDBSCAN's NMI on L2, cosine, $\chi^2$, and JS on Mnist8m with $k = 10, m = 100$.}
	\label{fig:sDbscan-Mnist8m-100}
\end{figure*} 

\subsection{Mnist8m}

Figure~\ref{fig:sOptics-Mnist8m-100} shows sOPTICS graphs on Mnist8m with L2, cosine, $\chi^2$ and JS distances where we use $\sigma = 4000$ for L2, and $d' = d$ on $\chi^2$ and JS.
It shows the advantages of sDBSCAN on supporting many distances and predicts that cosine, $\chi^2$ and JS provide higher accuracy than L2.

Figure~\ref{fig:sDbscan-Mnist8m-100} shows the accuracy of sDBSCAN variants compared to other competitors on Mnist8m with studied distances over the range of $\eps$ suggested by their corresponding sOPTICS graphs.
We consider \textit{sDBSCAN-1NN} with the approximation 1NN heuristic to cluster border and noisy points detected by sDBSCAN.
It shows again that sDBSCAN variants outperform sngDBSCAN on all studied distances.
sDBSCAN-1NN on a single computer nearly reaches the accuracy of kernel $k$-means~\cite{KernelKmean_Nys}.
Regarding the speed, each instance of sDNSCAN-1NN runs in less than 15 minutes, which is the time requirement for kernel $k$-means on a supercomputer.


\end{document}